\PassOptionsToPackage{frozencache}{minted}

\IfFileExists{./prepreamble-amspreprint.sty}{\RequirePackage[packages,theorems,changes]{prepreamble-amspreprint}}{}

\makeatletter
\providecommand*{\input@path}{}\edef\input@path{{./scoop-latex}\input@path}
\makeatother
\PassOptionsToPackage{style = authoryear-comp, backend = biber, dashed = false}{biblatex}
\PassOptionsToPackage{commentmarkup = footnote}{changes}
\RequirePackage{algorithm}
\RequirePackage{algpseudocode}    

\documentclass[english]{amsart}

\RequirePackage{biblatex}
\RequirePackage{ltxcmds}
\IfFileExists{./preamble-amspreprint.sty}{\RequirePackage[packages,theorems,changes]{preamble-amspreprint}}{}

\usepackage{scoop-local}

\IfFileExists{./postpreamble-amspreprint.sty}{\RequirePackage[packages,theorems,changes]{postpreamble-amspreprint}}{}

\makeatletter
\@ifpackageloaded{changes}{
\definechangesauthor[name = {Manuel Weiß}, color = {red!80!black}]{MW}
\definechangesauthor[name = {Lukas Baumgärtner}, color = {blue!80!black}]{LB}
\definechangesauthor[name = {Laura Weigl}, color = {orange!80!black}]{LW}
\definechangesauthor[name = {Ronny Bergmann}, color = {green!70!black}]{RB}
\definechangesauthor[name = {Stephan Schmidt}, color = {teal}]{STS}
\definechangesauthor[name = {Roland Herzog}, color = {TolVibrantBlue}]{RH}
}{}
\makeatother        

\makeatletter
\@ifpackageloaded{hyperref}{%
	\hypersetup{
		pdftitle = {Two Models for Surface Segmentation using the Total Variation of the Normal Vector},
		pdfauthor = {Manuel Weiß, Lukas Baumgärtner, Laura Weigl, Ronny Bergmann, Stephan Schmidt, Roland Herzog},
		pdfkeywords = {total variation, surface segmentation, non-smooth optimization, ADMM, Newton's method},
		pdfcreator = {Created using the Scoop Template Engine version 1.6.0.}
	}
}{
	\pdfinfo{
		/Title (Two Models for Surface Segmentation using the Total Variation of the Normal Vector)
		/Author (Manuel Weiß, Lukas Baumgärtner, Laura Weigl, Ronny Bergmann, Stephan Schmidt, Roland Herzog)
		/Subject ()
		/Keywords (total variation, surface segmentation, non-smooth optimization, ADMM, Newton's method)
		/Creator (Created using the Scoop Template Engine version 1.6.0.)
	}
}
\makeatother

\title[Surface Segmentation using the Total Variation of the Normal]{Two Models for Surface Segmentation using the Total Variation of the Normal Vector}

\author[M. Weiß]{Manuel Weiß\orcidlink{0000-0002-6098-9725}}
\address[M. Weiß]{Interdisciplinary Center for Scientific Computing, Heidelberg University, 69120 Heidelberg, Germany}
\email{\detokenize{manuel.weiss@iwr.uni-heidelberg.de}}
\urladdr{https://scoop.iwr.uni-heidelberg.de}

\author[L. Baumgärtner]{Lukas Baumgärtner\orcidlink{0000-0003-1007-4815}}
\address[L. Baumgärtner]{Institut für Mathematik, Humboldt University of Berlin, 10099 Berlin, Germany}
\email{\detokenize{lukas.baumgaertner@hu-berlin.de}}
\urladdr{https://www.mathematik.hu-berlin.de/en/people/mem-vz/1693318}

\author[L. Weigl]{Laura Weigl\orcidlink{0009-0006-4520-5968}}
\address[L. Weigl]{Universität Bayreuth, Universitätsstraße 30, 95447 Bayreuth, Germany}
\email{\detokenize{laura.weigl@uni-bayreuth.de}}
\urladdr{https://num.math.uni-bayreuth.de/de/team/laura-weigl}

\author[R. Bergmann]{Ronny Bergmann\orcidlink{0000-0001-8342-7218}}
\address[R. Bergmann]{Norwegian University of Science and Technology, Department of Mathematical Sciences, NO-7041 Trondheim, Norway}
\email{\detokenize{ronny.bergmannn@ntnu.no}}
\urladdr{https://www.ntnu.edu/employees/ronny.bergmann}

\author[S. Schmidt]{Stephan Schmidt\orcidlink{0000-0002-4888-0794}}
\address[S. Schmidt]{University of Trier, Universitätsring 15, 54296 Trier, Germany}
\email{\detokenize{s.schmidt@uni-trier.de}}
\urladdr{https://www.math.uni-trier.de/\string~schmidt}

\author[R. Herzog]{Roland Herzog\orcidlink{0000-0003-2164-6575}}
\address[R. Herzog]{Interdisciplinary Center for Scientific Computing, Heidelberg University, 69120 Heidelberg, Germany}
\email{\detokenize{roland.herzog@iwr.uni-heidelberg.de}}
\urladdr{https://scoop.iwr.uni-heidelberg.de}

\thanks{This work was supported by DFG grants HE~6077/10--2 and SCHM~3248/2--2 within the Priority Program SPP~1962 (Non-smooth and Complementarity-based Distributed Parameter Systems: Simulation and Hierarchical Optimization), which is gratefully acknowledged.}

\date{}

\dedicatory{}

\begin{document}

\begin{abstract}
We consider the problem of surface segmentation, where the goal is to partition a surface represented by a triangular mesh.
The segmentation is based on the similarity of the normal vector field to a given set of label vectors.
We propose a variational approach and compare two different regularizers, both based on a total variation measure.
The first regularizer penalizes the total variation of the assignment function directly, while the second regularizer penalizes the total variation in the label space.
In order to solve the resulting optimization problems, we use variations of the split Bregman (ADMM) iteration adapted to the problem at hand.
While computationally more expensive, the second regularizer yields better results in our experiments.
In particular it removes noise more reliably in regions of constant curvature.
In order to mitigate the computational cost, we present a manifold Newton scheme for the most expensive subproblem, which is related to the Riemannian center of mass on a sphere.
This significantly improves the computational cost.

\end{abstract}


\makeatletter
\ltx@ifpackageloaded{hyperref}{%
\subjclass[2010]{\href{https://mathscinet.ams.org/msc/msc2020.html?t=65D18}{65D18}, \href{https://mathscinet.ams.org/msc/msc2020.html?t=68U10}{68U10}, \href{https://mathscinet.ams.org/msc/msc2020.html?t=49M29}{49M29}, \href{https://mathscinet.ams.org/msc/msc2020.html?t=65K05}{65K05}, \href{https://mathscinet.ams.org/msc/msc2020.html?t=90C30}{90C30}}
}{%
}
\makeatother

\maketitle

\section{Introduction}
\label{section:introduction}

Segmentation of surfaces is a fundamental task in computer vision and shape analysis.
Its goal is to partition a given surface into disjoint regions based on certain features.
In this paper, we consider surfaces that are represented by a mesh~$\mesh$ consisting of a set~$\triangles$ of triangles and a set~$\edges$ of edges in $\R^3$.
As the feature, we employ the unit outer normal vector field~$\bn$ of the surface, which is piecewise constant.
Segmentation then becomes the task of assigning a label to each triangle~$\triangle \in \triangles$, based on a measure of similarity of the triangle's normal vector~$\bn_\triangle$ to a set of prescribed label vectors $\labels_1, \ldots, \labels_L$ belonging to the unit sphere $\Sphere \coloneqq \setDef{\bx \in \R^3}{\abs{x}_2 = 1}$.

The result of a segmentation is expressed in terms of an assignment function $\assign \colon \triangles \to \Simplex$.
Its value $\assign_\triangle$ on each triangle~$\triangle$ belongs to the probability simplex $\Simplex \coloneqq \setDef{\bx \in \R^L}{\bone^\transp \bx = 1, \; \bx \ge \bnull}$.
The label that will eventually be assigned to a triangle~$\triangle$ is the one determined by the dominant entry in~$\assign_\triangle$.

We follow a classical variational approach by minimizing a fidelity term that measures the similarity~$\labelindex{\similarity}(\bn)$ of the normal vector$~\bn$ to any of the label vectors~$\gl$.
We employ the geodesic distance~$\metricSymbol$ on the sphere~$\Sphere$ for this purpose.
That is, we have $\labelindex{\similarity}(\bn) \coloneqq \metric{\bn}{\gl}$, which amounts to the angle between $\bn$ and $\gl$.
These values can be pre-computed.
We obtain the problem
\begin{equation}
	\label{eq:general-model}
	\underset{\assign \colon \triangles \to \Simplex}{\text{Minimize}}
	\quad
	\sum_{\triangle \in \triangles} \abs{\triangle}
	\sum_{\ell=1}^L \labelindex{\similarity}(\bn_\triangle) \, \doubleindex{\assign}
	+
	\TVweight \, \cR(\assign)
	.
\end{equation}
In \eqref{eq:general-model}, $\abs{\triangle}$ denotes the area of~$\triangle$, $\TVweight > 0$ is a regularization parameter and $\cR$ is a regularizing functional.

The goal of this paper is to compare two regularization approaches, one of which is new in surface segmentation problems.
The first regularizer was originally proposed for image segmentation problems in \cite{LellmannKappesYuanBeckerSchnoerr:2009:1} and further developed in \cite{LellmannSchnoerr:2011:1,HeYousuffHussainiMaShafeiSteidl:2012:1}.
It measures the total variation of the assignment function $\assign$ in the assignment simplex~$\Simplex$, and thus we refer to it as \emph{assignment space total variation}, or A-TV in short.
The variational problem \eqref{eq:general-model} at hand has the same structure as in \cite{LellmannKappesYuanBeckerSchnoerr:2009:1}, but the pixels of the image domain are replaced by the triangles of the surface mesh and their color or gray-scale values are replaced by the normal vector data.

A drawback of A-TV is that it does not incorporate any metric structure in the space of labels.
Any transition of the labels assigned to neighboring triangles is penalized by equal amounts, regardless of the distance between the labels.
For the purpose of comparison with A-TV, we therefore propose an alternative regularizer, which measures the total variation in the label space~$\Sphere$ instead.
This means that incremental variations of the normal vector along a geodesic path on the sphere will by penalized no more than a single jump between the end points of that path.
We refer to this new approach as \emph{label space total variation}, or L-TV in short.

We can expect that L-TV will allow for smoother assignments, \eg, in regions of constant curvature, where neighboring normal vectors are close to each other.
On the other hand, we acknowledge that L-TV is more involved computationally.
This is primarily due to the fact that any value of the assignment function $\assign_\triangle$ that is not a vertex of the simplex~$\Simplex$ corresponds to a mixture of labels.
Due to the nonlinearity of the sphere~$\Sphere$ as a space, this mixture is not simply a convex combination but rather a nonlinear weighted average known as the Riemannian center of mass; see \cite{Karcher:1977:1}.

The Riemannian center of mass problem is a prominent optimization problem on Riemannian manifolds.
A variant of it appears as a subproblem in the algorithm we propose to minimize the L-TV model.
In order to mitigate the computational cost of this expensive step, we present a novel Newton scheme based on \cite{WeiglSchiela:2024:1} that solves these subproblems more efficiently than gradient descent in our setting.

\subsection*{Related Literature}

We briefly discuss alternative approaches to surface segmentation different from A-TV and L-TV.
In \cite{WuZhangDuanTai:2012:1}, the authors consider a variational model similar to A-TV but with the squared distance as similarity measure $\labelindex{\similarity}(\bn) \coloneqq \metric{\bn}{\gl}^2$.
In \cite{LellmannSchnoerr:2011:1}, the authors use so-called interaction potentials to achieve a similar effect as the label space total-variation (L-TV) regularizer studied below.
The interaction potentials allow for different penalization of label-jumps.
The method proposed in \cite{HwanKimDongYunUkLee:2006:1,WangYu:2011:1} iteratively merges triangles into larger regions, also using the outer normal vector field as a feature.
Furthermore, \cite{SunHarikBaek:2018:1,GauthierPuechBeniereSubsol:2017:1,YamauchiGumholdZayerSeidel:2005:1} compute an assignment based on discrete notions of mesh curvature, derived from the normal vector~$\bn$.
In an alternative variational approach, \cite{CohenSteinerAlliezDesbrun:2004:1,YanWangLiuYang:2012:1} achieve segmentation by minimizing the distance of the given mesh to an idealized mesh associated with the segmentation.
The authors in \cite{NabiDouik:2016:1,ZhangZhengWuCai:2012:1,ZhangWuDengLiuYang:2017:1} propose a segmentation approach based on minimizing the Mumford-Shah functional \cite{MumfordShah:1989:1}, which is well-known from imaging applications.
Finally, there is a stream of literature in geometric and topological data science, in which neural networks are trained to perform the task of surface segmentation or related inverse shape problems; see, \eg, \cite{HanockaHertzFishGiryesFleishmanCohenOr:2019:1,CharlesSuKaichunGuibas:2017:1}.

\subsection*{Contributions}

We investigate two total-variation regularizers for surface segmentation problems based on the normal vector field as a feature.
The \emph{assignment space total variation} regularizer (A-TV) is a straightforward adaptation from image to surface segmentation.
The \emph{label space total variation} (L-TV) also borrows ideas previously used for manifold-valued imaging segmentation but appears to be new in the context of surface segmentation.
It takes into account the metric structure of the label space, \ie, the unit sphere since we use the normal vector as feature.
We develop an Alternating Direction Method of Multipliers (ADMM) approach to solve L-TV problems.
In particular, we present a novel Newton scheme --- an algorithm that may be of independent interest --- for problems closely related to the Riemannian center of mass arising in every ADMM iteration.
We compare the segmentation results obtained by A-TV and L-TV in numerical experiments that demonstrate that the L-TV model may yield smoother assignment functions, albeit at higher computational cost.

\subsection*{Outline}

This paper is structured as follows:
\Cref{section:prelminiaries} introduces the required notions on triangulated surfaces and differential geometry of the sphere~$\Sphere$ and the probability simplex~$\Simplex$.
In \cref{section:two-total-variation-regularizers}, we define the A-TV and L-TV models and discuss their differences.
\Cref{section:numerical-algorithms} presents algorithms to solve each problem numerically.
This specifically includes a new manifold Newton scheme for one of the subproblems of the L-TV model that is related to the Riemannian center of mass.
\Cref{section:numerical-examples} presents numerical results that allow A-TV to be compared to L-TV.

\section{Preliminaries}
\label{section:prelminiaries}

\subsection{Discrete Surfaces}
\label{subsection:discrete-surfaces}

In this work, we consider triangulated surface meshes $\mesh$ that are embedded in $\R^3$.
We denote the set of triangles by~$\triangles$ and the set of edges by~$\edges$.
We work with manifold meshes without boundary, \ie, we assume that every edge $\edge$ is connected to exactly two triangles.
The two sides of an edge are denoted by $\edge_+$ and $\edge_-$, respectively, with arbitrary but fixed orientation.
Furthermore, we assume that the surface is oriented, \ie, there exists a global outer unit normal vector field~$\bn$.

On the mesh~$\mesh$, we define the space of piecewise constant, $X$-valued functions
\begin{equation*}
	\DG{\triangles}{X}
	\coloneqq
	\setDef[big]
	{\bu \colon \triangles \to X}
	{\restr{\bu}{\triangle} \in P_0(\triangle,X) \text{ for all } \triangle \in \triangles}
	.
\end{equation*}
Here, $P_0(\triangle,X)$ denotes the space of constant functions on the triangle~$\triangle$ with values in~$X$.
For instance, the fact that the unit normal vector field~$\bn$ is piecewise constant and has values on the sphere can be expressed as $\bn \in \DG{\triangles}{\Sphere}$.

On the skeleton (the union of edges), we analogously define
\begin{equation*}
	\DG{\edges}{X}
	\coloneqq
	\setDef[big]
	{\bu \colon \edges \to X}
	{\restr{\bu}{\edge} \in P_0(\edge,X) \text{ for all } \edge \in \edges}
	.
\end{equation*}
We denote the constant value of $\bu$ on a triangle $\triangle$ by $\bu_\triangle \coloneqq \restr{\bu}{\triangle}$ for $\bu \in \DG{\triangles}{X}$.
Analogously, we write $\bu_\edge \coloneqq \restr{\bu}{\edge}$ for $\bu \in \DG{\edges}{X}$.

\subsection{Differential Geometry on the Sphere}
\label{subsection:differential-geometry-sphere}

Given a point $\bm \in \cS$, we denote its tangent plane by $\tangent{\bm}[\Sphere]$.
The tangent bundle $\tangentBundle[\Sphere]$ is the disjoint union of all tangent planes, endowed with the smooth structure inherited from~$\Sphere$.
We choose the Euclidean inner product from the embedding $\tangent{\bm}[\Sphere] \subseteq \R^3$ as the Riemannian metric on $\Sphere$, turning the sphere into a Riemannian manifold.
We denote the norm of a tangent vector by $\abs{\, \cdot \,}_2$.
The exponential map describes the point on the sphere reached by following the geodesic starting at~$\bm \in \cS$ in tangential direction~$\bX$ for unit time.
It is explicitly given by
\begin{equation*}
	\exponential{\bm}
	\colon
	\tangent{\bm}[\Sphere]
	\to
	\cS
	,
	\quad
	\bX
	\mapsto
	\exponential{\bm} \bX
	=
	\cos\paren(){\abs{\bX}_2} \, \bm
	+
	\sin\paren(){\abs{\bX}_2}
	\frac{\bX}{\abs{\bX}_2}
	,
\end{equation*}
see for instance the appendix of \cite{BergmannHerrmannHerzogSchmidtVidalNunez:2020:1} or \cite[Exercise~10.38]{Boumal:2023:1}.
The inverse mapping $\logarithm{\bm}$ is well-defined except on pairs of antipodal points.
It satisfies
\begin{equation*}
	\exponential[big]{\bm}(\logarithm{\bm}(\widetilde{\bm}))
	=
	\widetilde{\bm}
	\quad
	\text{for }
	\widetilde{\bm} \neq - \bm
	.
\end{equation*}
The parallel transport $\parallelTransport{\bm}{\widetilde{\bm}} \colon \tangent{\bm}[\Sphere] \to \tangent{\widetilde{\bm}}[\Sphere]$ transports a tangent vector at $\bm$ to a tangent vector $\bX \in \tangent{\bm}[\Sphere]$ at $\widetilde{\bm}$ along the unique shortest geodesic from~$\bm$ to $\widetilde{\bm} \neq - \bm$.
The parallel transport has the following explicit representation,
\begin{equation}
	\parallelTransport{\bm}{\widetilde{\bm}}(\bX)
	=
	\bX
	-
	2
	\,
	\frac{\bX^\transp (\bm + \widetilde{\bm})}{\abs{\bm + \widetilde{\bm}}_2^2}
	\,
	(\bm + \widetilde{\bm})
	,
	\label{eq:parallel-transport}
\end{equation}
where $\bm, \widetilde{\bm} \in \Sphere$ and again $\widetilde{\bm} \neq -\bm$ is required.

The choice of the Riemannian metric, \ie, the family of inner products in the tangent spaces $\tangent{\bm}[\Sphere]$, induces a notion of distance on~$\Sphere$.
For two points $\widetilde{\bm}, \bm \in \Sphere$, the geodesic distance agrees with the \enquote{great arc distance}
\begin{equation}
	\label{eq:sphere:distance}
	\metric{\widetilde{\bm}}{\bm}[\Sphere]
	=
	\arccos \paren(){\widetilde{\bm} \cdot \bm}
	=
	\sphericalangle(\widetilde{\bm}, \bm)
	,
\end{equation}
where $\sphericalangle(\widetilde{\bm}, \bm)$ denotes the angle (arc-length).
As on any Riemannian manifold, the geodesic distance is related to the logarithmic map via
\begin{equation}
	\label{eq:sphere:distance-via-log}
	\metric{\widetilde{\bm}}{\bm}[\Sphere]
	=
	\abs{\logarithm{\bm}(\widetilde{\bm})}_2
	,
\end{equation}
whenever the latter is defined; see also~\cite{GotoSato:2021:1}.
The logarithmic map on the sphere $\Sphere$ is given by
\begin{equation}
	\label{eq:sphere:logarithmic-map}
	\logarithm{\bm}(\widetilde{\bm})
	=
	\begin{cases}
		\mathbf{0}
		&
		\text{if }
		\widetilde{\bm} = \bm
		,
		\\
		\arccos \paren(){\widetilde{\bm} \cdot \bm} \, \frac{\widetilde{\bm} - (\bm \cdot \widetilde{\bm}) \, \bm}{\abs{\widetilde{\bm} - (\bm \cdot \widetilde{\bm}) \, \bm}_2}
		&
		\text{otherwise}
		,
	\end{cases}
\end{equation}
where again the case $\widetilde{\bm}=- \bm$ is excluded.
It is worth noting that using~\eqref{eq:sphere:distance-via-log} over~\eqref{eq:sphere:distance} provides considerably more information about the structure of the non-smoothness of the problem.
It is well known that $\widetilde{\bm} \mapsto \logarithm{\bm}(\widetilde{\bm})$ is a smooth function in particular at $\widetilde{\bm} = \bm$.
Hence the non-smoothness of~\eqref{eq:sphere:distance} stems solely from the non-differentiability of the norm $\abs{\, \cdot \,}_2$ at the tangent's space origin.

For the Newton scheme we devise in \cref{subsubsection:ltv-problem:m-subproblem}, we will need second derivatives of the logarithmic map~\eqref{eq:sphere:logarithmic-map}.
To this end, we state the following result whose proof is straightforward.
We rewrite $\logarithm{\bm}(\widetilde{\bm}) = h(\widetilde{\bm}\cdot \bm) \, (\widetilde{\bm} - (\bm \cdot \widetilde{\bm}) \, \bm)$, where

\begin{equation}
	\label{eq:log-derivative:h-definition}
	h \colon \interval(]{-1}{1} \to \R
	,
	\quad
	h(x)
	=
	\begin{cases}
		\frac{\arccos(x)}{\sqrt{1 - x^2}}
		&
		\text{if }
		x < 1
		,
		\\
		1
		,
		&
		\text{else}
		.
	\end{cases}
\end{equation}
The derivatives of $h$ for $x \in \interval(]{-1}{1}$ are given by
\begin{subequations}
  \label{eq:log-derivative:h-derivatives}
	\begin{align}
		\label{eq:log-derivative:h-derivatives:first}
		h'(x)
		&
		=
		\frac{1}{\paren(){1-x^2}^{3/2}}
		\paren[big](){%
			x \arccos(x)
			-
			\sqrt{1-x^2}
		}
		,
		\\
		\label{eq:log-derivative:h-derivatives:second}
		h''(x)
		&
		=
		\frac{1}{\paren(){1-x^2}^{5/2}}
		\paren[big](){%
			\arccos(x) \paren(){1+2x^2}
			-
			3x \sqrt{1-x^2}
		}
		.
	\end{align}
\end{subequations}
In particular, we have
\begin{equation}
	\label{eq:log-derivative:limits}
	\lim_{x \to 1^-}
	h(x)
	=
	1
	,
	\quad
	\lim_{x \to 1^-}
	h'(x)
	=
	-\frac{1}{3}
	,
	\quad
	\lim_{x \to 1^-}
	h''(x)
	=
	\frac{4}{15}
	.
\end{equation}
We further recall the \emph{Riemannian center of mass}, a generalization of the Euclidean mean on manifolds, as proposed by \cite{Karcher:1977:1}.
Given $L$~points $\gl \in \Sphere$, $\ell = 1, \ldots, L$ and corresponding weights $\assign_\ell \in \R_{\ge 0}$, a point $\bm \in \Sphere$ is said to be a Riemannian center of mass \wrt the data $(\assign_\ell,\gl)$ if
\begin{equation}
	\label{eq:Karcher-mean:definition}
	\bm
	\in
	\argmin_{\bm \in \Sphere}
	\sum_{\ell=1}^L \labelindex{\assign} \, \metric{\bm}{\gl}[\Sphere]^2
	.
\end{equation}
In general, there is no closed-form solution to compute such~$\bm$.
However, if a Riemannian center of mass $\bm$ fulfills $\bm \neq - \gl$ for all $\ell = 1, \ldots, L$, the necessary optimality condition reads
\begin{equation}
	\label{eq:Karcher-mean:optimality-condition}
	0
	=
	\sum_{\ell=1}^L \assign_\ell \logarithm{\bm}(\gl)
	.
\end{equation}
Notice that in Euclidean space, rather than the sphere, $\logarithm{\bm}(\gl) = \gl - \bm$ holds and thus \eqref{eq:Karcher-mean:optimality-condition} amounts to $\bm = (\sum_{\ell=1}^L \assign_\ell \, \gl)/(\sum_{\ell=1}^L \assign_\ell)$, thus explaining why the Riemannian center of mass generalizes the Euclidean mean.

For $\bm \in \DG{\triangles}{\Sphere}$, we further define
\begin{align*}
	\DGat{\triangles}{\tangentBundle[\Sphere]}{\bm}
	&
	\coloneqq
	\setDef[big]
	{\bY \in \DG{\triangles}{\tangentBundle[\Sphere]}}
	{\bY_\triangle \in \tangent{\bm_\triangle}[\Sphere] \text{ for all } \triangle \in \triangles}
	,
	\\
	\DGat{\edges}{\tangentBundle[\Sphere]}{\bm}
	&
	\coloneqq
	\setDef[big]
	{\bX \in \DG{\edges}{\tangentBundle[\Sphere]}}
	{\bX_\edge \in \tangent{\eplus{\bm}}[\!\!\Sphere] \text{ for all } \edge \in \edges}
	.
\end{align*}

\subsection{Differential Geometry on the Simplex}
\label{subsection:differential-geometry-simplex}

In addition to the closed probability simplex
\begin{equation*}
	\Simplex
	=
	\setDef[big]{\assign \in \R^L}{\bone^\transp \assign = 1, \; \assign \ge \bnull}
	,
\end{equation*}
we also require the open probability simplex
\begin{equation*}
	\opensimplex
	\coloneqq
	\setDef[big]{\assign \in \R^L}{\bone^\transp \assign = 1, \; \assign > \bnull}
	.
\end{equation*}
The tangent space $\tangent{\assign}[\opensimplex]$ at any point $\assign \in \opensimplex$ is given by
\begin{equation*}
	\tangent{\assign}[\opensimplex]
	=
	\setDef[big]{\bX \in \R^L}{\bone^\transp \bX = 0}
	.
\end{equation*}
Using the Fisher-Rao metric, $\opensimplex$ can be equipped with a Riemannian manifold structure; see, \eg, \cite[Sec.~2.1]{AastroemPetraSchmitzerSchnoerr:2017:1} for details.
The geodesic distance increases compared to the Euclidean distance close to the boundary.
Given a point $\assign \in \opensimplex$, the exponential map $\exponential{\assign} \colon \tangent{\assign}[\opensimplex] \to \opensimplex$ is given by
\begin{equation}
  \label{eq:simplex:exponential-map}
	\exponential{\assign}(\bX)
	=
	\frac{1}{2}
	\paren[bigg](){\assign + \frac{\Xphi^2}{\abs{\Xphi}^2}}
	+
	\frac{1}{2}
	\paren[bigg](){\assign - \frac{\Xphi^2}{\abs{\Xphi}^2}}
	\cos \paren(){\abs{\Xphi}}
	+
	\frac{\sin \paren(){\abs{\Xphi}}}{\abs{\Xphi}}
	\Xphi \odot \sqrt{\assign}
	,
\end{equation}
where $\Xphi = \bX / \sqrt{\assign}$ and $\odot,/, \sqrt{}$ are meant element-wise.
Given a function $f \colon \R^L \to \R$ and a point $\assign \in \opensimplex$, the Riemannian gradient $\gradsimplex f(\assign)$ at $\assign$ can be computed using the Euclidean gradient $\nabla f(\assign)$ by
\begin{equation}
  \label{eq:simplex:Riemannian-gradient}
  \gradsimplex f (\assign)
  =
  \nabla f(\assign) \odot \assign
  -
  \paren[big](){\assign^\transp \nabla f(\assign)} \, \assign
  .
\end{equation}

\section{Segmentation Using Two Total Variation Regularizers}
\label{section:two-total-variation-regularizers}

In this section, we discuss two regularizes~$\cR$ for the segmentation problem \eqref{eq:general-model}: the \emph{assignment space total variation} (A-TV) and the \emph{label space total variation} (L-TV).
We recall from \eqref{eq:general-model} the common framework is to find a piecewise constant assignment function $\assign \in \DG{\triangles}{\Simplex}$ that minimizes
\begin{equation*}
	\underset{\assign \in \DG{\triangles}{\Simplex}}{\text{Minimize}}
	\quad
	\sum_{\triangle \in \triangles} \abs{\triangle}
	\sum_{\ell=1}^L \labelindex{\similarity}(\bn_\triangle) \, \doubleindex{\assign}
	+
	\TVweight \, \cR(\assign)
	.
\end{equation*}
In the absence of a regularizer ($\TVweight = 0$), the problem decouples and the minimum of $\labelindex{\similarity}(\bn_\triangle) \, \doubleindex{\assign}$ is attained by the $\ell$-th unit vector
that corresponds to the label vector in $\set{\labels_1, \ldots, \labels_L}$ that is closest to $\bn_\triangle$.

The purpose of the regularizer~$\cR$ is to counteract noise in the normal vector data and achieve clean segmentation results.
Both regularizers are based on the total variation for piecewise constant functions $\bu \in \DG{\triangles}{X}$ with values in a metric space $(X,\metricSymbol_X)$, defined as \cite{BergmannHerrmannHerzogSchmidtVidalNunez:2020:2, LellmannStrekalovskiyKoetterCremers:2013:1}
\begin{equation}
	\label{eq:metric-space:total-variation}
	\TV_X(\bu)
	\coloneqq
	\sum_{\edge \in \edges} \abs{\edge} \, \metric{\eplus{\bu}}{\eminus{\bu}}[X]
	,
\end{equation}
where $\abs{\edge}$ is the length of the edge~$\edge$.

\subsection{Segmentation Using Regularization by Assignment Space Total Variation}
\label{subsection:assignment-space-tv}

In a straightforward adaptation of \cite{LellmannKappesYuanBeckerSchnoerr:2009:1}, we directly penalize the total variation of the assignment function~$\assign$.
This corresponds to choosing $X = \Simplex$ and $\bu = \assign$ and it leads to the \emph{assignment space total variation} problem
\begin{equation}
	\label{eq:atv:model-problem}
	\tag{A-TV}
	\underset{\assign \in \DG{\triangles}{\Simplex}}{\text{Minimize}}
	\quad
	\sum_{\triangle \in \triangles} \abs{\triangle}
	\sum_{\ell=1}^L \labelindex{\similarity}(\bn_\triangle) \, \doubleindex{\assign}
	+
	\TVweight \, \assignmentTV(\assign)
	.
\end{equation}
We equip the simplex~$\Simplex$ with the metric induced by the $1$-norm $\abs{\, \cdot \,}_1$, so that the total variation term~$\assignmentTV$ according to \eqref{eq:metric-space:total-variation} becomes
\begin{equation}
	\label{eq:atv:total-variation}
	\assignmentTV(\assign)
	=
  \sum_{\edge \in \edges} \abs{\edge} \, \abs{\eplus{\assign} - \eminus{\assign}}_1
	.
\end{equation}

\subsection{Segmentation Using Regularization by Label Space Total Variation}
\label{subsection:label-space-tv}

As an alternative to \eqref{eq:atv:model-problem}, we propose a second regularizer that is based on the total variation in the label space~$\Sphere$.
Taking into account that the value of the assignment function~$\assign$ in the simplex~$\Simplex$ represents a mixture of labels, we associate with $\assign$ a corresponding element $\mphi$ of the unit sphere~$\Sphere$.
Since the latter is a nonlinear space, a mixture of these labels is not simply a convex combination but rather a nonlinear weighted average, the Riemannian center of mass \eqref{eq:Karcher-mean:definition}.
On the triangle~$\triangle$, this nonlinear mixture of the labels $\labels_1, \ldots, \labels_L$ is given by
\begin{equation}
	\label{eq:sphere:Karcher-mean-problem}
	\mphi_\triangle
	\in
	\argmin_{\bm \in \Sphere}
	\sum_{\ell=1}^L \doubleindex{\assign} \, \metric{\bm}{\gl}[\Sphere]^2
	,
\end{equation}
so that $\mphi$ is a piecewise constant function $\mphi \in \DG{\triangles}{\Sphere}$.
In terms of \eqref{eq:metric-space:total-variation}, the \emph{label space total variation} problem corresponds to the choice $X = \Sphere$ and $\bu = \mphi$ and it reads
\begin{equation}
	\label{eq:ltv:model-problem}
	\tag{L-TV}
	\underset{\assign \in \DG{\triangles}{\Simplex}}{\text{Minimize}}
	\quad
	\sum_{\triangle \in \triangles} \abs{\triangle}
	\sum_{\ell=1}^L \labelindex{\similarity}(\bn_\triangle) \, \doubleindex{\assign}
	+
	\TVweight \, \labelTV[\mphi]
	.
\end{equation}
In contrast to \eqref{eq:atv:total-variation}, the total variation term now considers the geodesic distances of neighboring Riemannian centers of mass,
\begin{equation}
	\label{eq:ltv:total-variation}
	\labelTV[\mphi]
	=
	\sum_{\edge \in \edges} \abs{\edge} \, \metric[big]{\eplus{\mphi}}{\eminus{\mphi}}[\Sphere]
	.
\end{equation}

\begin{figure}[htb]
	\centering
	\begin{subfigure}[b]{0.47\textwidth}
		\centering
		\includegraphics[width = \textwidth]{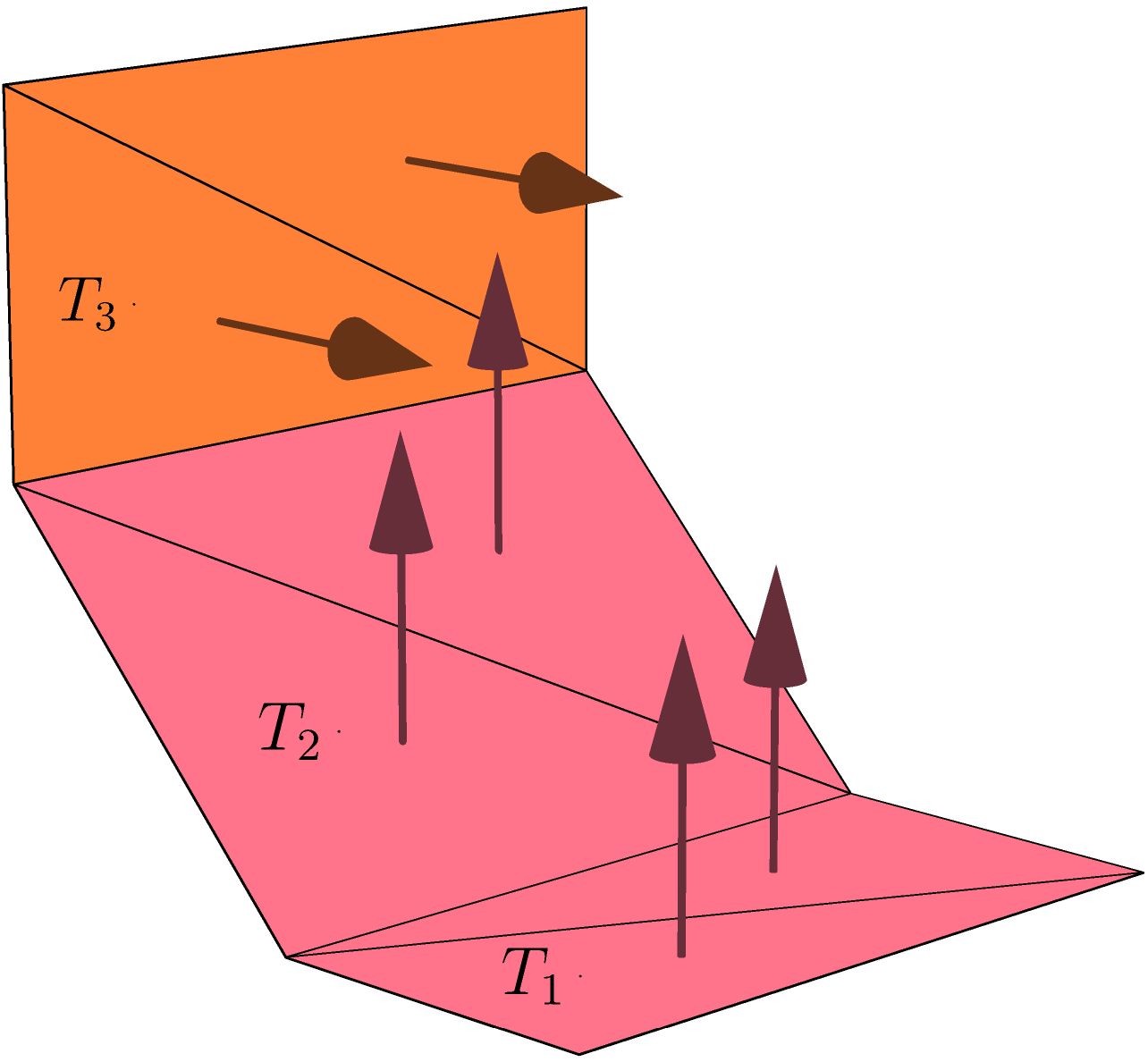}
		\caption{$\assignmentTV(\assign)=2$, $\labelTV[\mphi] = \frac{\pi}{2}$}
		\label{figure:comparison-of-the-regularizers:2-labels}
	\end{subfigure}
	\hfill
	\begin{subfigure}[b]{0.47\textwidth}
		\centering
		\includegraphics[width = \textwidth]{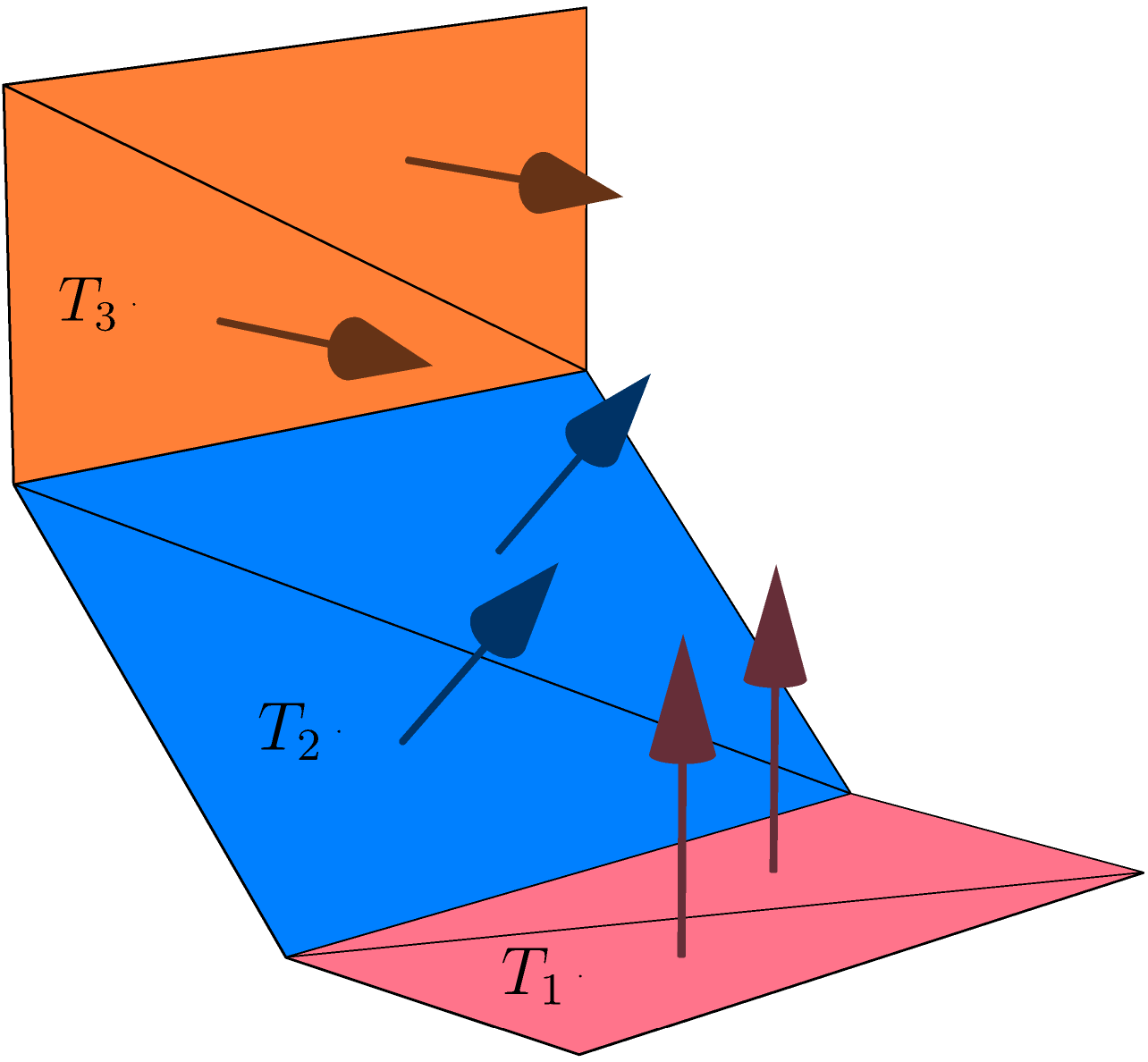}
		\caption{$\assignmentTV(\widetilde{\assign})=4$, $\labelTV[\mphi[\widetilde{\assign}]]= \frac{\pi}{2}$}
		\label{figure:comparison-of-the-regularizers:3-labels}
	\end{subfigure}
	\caption{Visualization of the assignments $\assign, \widetilde{\assign}$ as described in \cref{example:comparison-of-the-regularizers}.}
	\label{figure:comparison-of-the-regularizers}
\end{figure}

\begin{example}[Comparison of the regularizers]
	\label{example:comparison-of-the-regularizers}
	To emphasize the difference of the alternative model \eqref{eq:ltv:model-problem} compared to \eqref{eq:atv:model-problem}, we compare the values of $\assignmentTV[\assign]$ and $\labelTV[\mphi]$ for two different assignments~$\assign$ and $\widetilde{\assign}$.
	The situation is illustrated in \cref{figure:comparison-of-the-regularizers}.
	Normalizing the edges where jumps occur to length~one, we evaluate
	\begin{equation}
		\label{eq:comparison-of-the-regularizers}
		2
		=
		\assignmentTV[\assign]
		<
		\assignmentTV[\widetilde{\assign}]
		=
		4
		,
		\quad
		\labelTV[\mphi]
		=
		\labelTV[\mphi[\widetilde{\assign}]]
		=
		\frac{\pi}{2}
		.
	\end{equation}
	This clarifies that the assignment of an intermediate label between two assigned labels incurs an additional penalty in the assingment space total variation model \eqref{eq:atv:model-problem} but not in the \eqref{eq:ltv:model-problem} model.
	This hints at the fact later confirmed by our numerical experiments that the \eqref{eq:atv:model-problem} model tends to \enquote{skip} labels to reduce the regularization penalty.
\end{example}

\section{Numerical Algorithms}
\label{section:numerical-algorithms}

In this section, we present numerical methods for both models.
The simpler assignment space total variation problem \eqref{eq:atv:model-problem} can actually be rewritten as a linear program with $(\abs{\triangles} + \abs{\edges}) \cdot L$ variables, $2 L \cdot \abs{\edges}$ inequality constraints and $\abs{\triangles}$ equality constraints.
In our largest problem instance (\cref{subsection:numerical-examples:fandisk-mesh}),
this amounts to more than 1.4~million variables.
While problems of this size, particularly with sparse constraint matrices, can be handled by state-of-the-art solvers, in this study we use the Chambolle-Pock algorithm \cite{ChambollePock:2011:1,PockCremersBischofChambolle:2009:1}.
This leads to very simple and parallelizable update steps.

The Chambolle-Pock method cannot be directly applied to the label space assignment problem \eqref{eq:ltv:model-problem}, because the Riemannian center of mass~$\mphi$ does not depend linearly on the assignment~$\assign$.
We therefore use an alternating direction method of multipliers (ADMM) \cite{BoydParikhChuPeleatoEckstein:2010:1}, whose individual steps are also parallelizable and mostly simple.

\subsection{Chambolle-Pock Algorithm for Assignment Space Total Variation}
\label{subsection:atv-problem:Chambolle-Pock-algorithm}

For problem \eqref{eq:atv:model-problem}, we use the Chambolle-Pock algorithm \cite{PockCremersBischofChambolle:2009:1}, similarly as in \cite[Section~6.3.4]{ChambollePock:2011:1}.
To set up the algorithm, we introduce the jump operator~$K$
\begin{equation}
	\label{eq:atv:jump-operator}
	K
	\colon
	\DG{\triangles}{\Simplex}
	\to
	\DG{\edges}{\R^L}
	,
	\quad
	\paren[normal](){K \assign}_\edge
	\coloneqq
	\TVweight \, \paren(){\eplus{\assign} - \eminus{\assign}}
	.
\end{equation}
The algorithm can be be written as shown in \cref{algorithm:atv-problem:Chambolle-Pock-algorithm}, where $K^*$ denotes the Hilbert-space adjoint of $K$ \wrt the $L^2$-inner products on $\DG{\triangles}{\R^L}$ and $\DG{\edges}{\R^L}$.
These inner products are represented by diagonal matrices with the triangle areas or edge lengths on the diagonal, respectively.

\begin{algorithm}[htb]
	\caption{Chambolle-Pock algorithm applied to the assignment space total variation problem \eqref{eq:atv:model-problem}.}
	\label{algorithm:atv-problem:Chambolle-Pock-algorithm}
	\begin{algorithmic}[1]
		\Require
    similarity measure $\similarity \in \DG{\triangles}{\R^L}$
		\Require
		step sizes
		$\tau > 0$,
		$\sigma > 0$,
		extrapolation parameter
		$\theta \in [0,1]$
		\Require
		$\sequence{\assign}{0} \in \DG{\triangles}{\Simplex}$,
		$\sequence{\cpdual}{0} \in \DG{\edges}{\R^L}$
		\Ensure
		approximate solution of \eqref{eq:atv:model-problem}
		\State
		Set $\sequence{\overline{\assign}}{0} \gets \sequence{\assign}{0}$
		\While{not converged}
		\State
		Set
		$
		\sequence{\cpdual_\edge}{k+1}
		\gets
		\clip
		\paren[big](){
			\sequence{\cpdual_\edge}{k}
			+
			\sigma
			\paren[big][]{
				K \sequence{\overline{\assign}}{k}
			}_\edge
    ,-1,1
		}
		\quad
		\text{for all edges }
		\edge
		$
		\State
		Set
		$
		\sequence{\assign_\triangle}{k+1}
		\gets
		\proj_{\Simplex}
		\paren[big]()
		{
			\sequence{\assign_\triangle}{k}
			- \tau
			\paren[big][]{
        K^* \sequence{\cpdual}{k+1} - \tau s
			}_\triangle
		}
		\quad
		\text{for all triangles }
		\triangle
		$
		\label{step:atv-problem:Chambolle-Pock-algorithm:simplex-projection}
		\State
		Set
		$
		\sequence{\overline{\assign}}{k+1}
		\gets
		\sequence{\assign}{k+1}
		+
		\theta \,
		\paren(){
			\sequence{\assign}{k+1}
			-
			\sequence{\assign}{k}
		}
		$
		\State
		$k
		\gets
		k+1$
		\EndWhile
	\end{algorithmic}
\end{algorithm}

Here, the clipping operator $\clip(\cdot,-1,1) \colon \R^L \to[-1,1]^L$ projects each component to the interval $[-1,1]$.
Further, the simplex projection $\proj_{\Simplex}$ in \cref{step:atv-problem:Chambolle-Pock-algorithm:simplex-projection} can be efficiently computed using Algorithm~1 in \cite{WangCarreiraPerpinan:2013:1}.

We mention that \eqref{eq:atv:model-problem} can also be solved using other, potentially more efficient methods.
For instance, one can rewrite it as a min-cut linear program and apply algorithms for this problem class; see, for instance, \cite{YuanBaeTai:2010:1,BaeYuanTaiBoykov:2014:1,YuanBaeTaiBoykov:2010:1}.

\subsection{ADMM for Label Space Total Variation}
\label{subsection:ltv-problem:ADMM-algorithm}

\Cref{algorithm:atv-problem:Chambolle-Pock-algorithm} cannot be transferred easily to the label space assignment problem \eqref{eq:ltv:model-problem}, because the Riemannian center of mass~$\mphi$ does not depend linearly on the assignment~$\assign$.
A method applicable to \eqref{eq:ltv:model-problem} is the alternating direction method of multipliers (ADMM); see \cite{BoydParikhChuPeleatoEckstein:2010:1,GoldsteinOsher:2009:1}.

We now derive the specific form of that algorithm.
In addition to the assignment variable $\assign \in \DG{\triangles}{\Simplex}$, we introduce three auxiliary variables $\bm$, $\bY$, and $\bX$.
As is customary in ADMM, the purpose of these variables is to allow problem \eqref{eq:ltv:model-problem} to be split into simpler subproblems.
Each of the auxiliary variables is coupled to the remaining variables through a constraint that is enforced by a penalty term in the augmented Lagrangian function.

The first auxiliary variable $\bm \in \DG{\triangles}{\Sphere}$ represents the Riemannian center of mass of the labels, weighted according to the assignment function~$\assign$ per triangle.
We introduce the necessary optimality condition \eqref{eq:Karcher-mean:optimality-condition} of the weighted Riemannian center of mass problem as the corresponding constraint.
In fact, we replace the nonlinear terms $\logarithm{\bm}(\gl)$ in \eqref{eq:Karcher-mean:optimality-condition}, triangle by triangle, by the auxiliary variable $\bY \in \DG{\triangles}{\tangentBundle[\Sphere]}$.
This leaves us with the constraint \eqref{eq:Karcher-mean:optimality-condition} in the form $\sum_{\ell=1}^L \doubleindex{\assign} \doubleindex{\bY} = \bnull$ as well as $\doubleindex{\bY} = \logarithm{\bm_\triangle}(\gl)$.
For numerical stability, we represent the latter in the form $\gl = \exponential{\bm_\triangle} \doubleindex{\bY}$.

Finally, to handle the non-smoothness in the total variation \eqref{eq:ltv:total-variation}, we introduce the third auxiliary variable $\bX \in \DG{\edges}{\tangentBundle[\Sphere]}$ coupled by the constraint $\bX_\edge = \logarithm{\eplus{\bm}}(\eminus{\bm})$.
We thus arrive at
\begin{equation}
	\label{eq:ltv:model-problem:expanded}
	\begin{aligned}
		\underset{\assign, \bm, \bY, \bX}{\text{Minimize}}
		\quad
		&
		\sum_{\triangle \in \triangles}
		\abs{\triangle} \, \assign_\triangle^\transp \similarity_\triangle
		+
		\TVweight
		\sum_{\edge \in \edges} \abs{\edge} \abs{\bX_\edge}_2
		\\
		\text{subject to}
		\quad
		&
		\paren[auto]\{.{%
			\begin{aligned}
				&
				\bnull
				=
				\sum_{\ell=1}^L
				\doubleindex{\assign}
				\doubleindex{\bY}
				\in
				\tangent{\bm_\triangle}[\Sphere]
				\quad
				\text{for all }
				\triangle \in \triangles
				,
				\\
				&
				\gl
				=
				\exponential{\bm_\triangle} \doubleindex{\bY}
				\in
				\Sphere
				\subseteq
				\R^3
				\quad
				\text{for all }
				\triangle \in \triangles,
				\ell = 1, \ldots, L
				,
				\\
				&
				\bX_\edge
				=
				\logplusminus
				\in \tangent{\eplus{\bm}}[\!\!\Sphere]
				\quad
				\text{for all }
				\edge \in \edges
				.
			\end{aligned}
		}
	\end{aligned}
\end{equation}
Notice that \eqref{eq:ltv:model-problem:expanded} is not strictly equivalent to \eqref{eq:ltv:model-problem}, because we use a necessary, but not sufficient, optimality condition to substitute the Riemannian center of mass~$\bm$.
In practice, we did not observe any difficulties stemming from this discrepancy.

We associate with problem \eqref{eq:ltv:model-problem:expanded} the following augmented Lagrangian
\begin{align}
	\MoveEqLeft
	\cL_\rho
	\paren(){\assign, \bm, \bY, \bX, \dualkarcher, \dualextra, \dualjumps}
	\notag
	\\
	&
	=
	\sum_{\triangle \in \triangles}
	\abs{\triangle} \, \assign_\triangle^\transp \similarity_\triangle
	+
	\TVweight
	\sum_{\edge \in \edges}
	\abs{\edge} \abs{\bX_\edge}_2
	\notag
	+
	\frac{\rho}{2}
	\sum_{\triangle \in \triangles}
	\abs{\triangle}
	\abs[Big]{
		\sum_{\ell=1}^L
		\doubleindex{\assign} \doubleindex{\bY}
		+
		\dualkarcher_\triangle
	}_2^2
	\notag
	\\
	&
	\quad
	+
	\frac{\rho}{2}
	\sum_{\triangle \in \triangles}
	\abs{\triangle}
	\sum_{\ell=1}^L
	\abs[big]{
		\exponential{\bm_\triangle}(\doubleindex{\bY})
		-
		\gl
		+
		\doubleindex{\dualextra}
	}_2^2
	\notag
	\\
	&
	\quad
	+
	\frac{\rho}{2}
	\sum_{\edge \in \edges}
	\abs{\edge}
	\abs[big]{
		\logplusminus
		- \bX_\edge
		+ \dualjumps_\edge
	}_2^2
	.
	\label{eq:ltv:model-problem:Augmented-Lagrangian}
\end{align}
The quantities $\dualkarcher \in \DG{\triangles}{\R^3}$, $\dualextra \in \DG{\triangles}{\R^{L \times3}}$, as well as $\dualjumps \in \DG{\edges}{\R^3}$ represent the (rescaled) Lagrange multipliers associated with the constraints in \eqref{eq:ltv:model-problem:expanded}.
The values of $\bX$ and $\bY$ need to be constrained to belong to the respective tangent spaces of $\Sphere$.
Depending on~$\bm$, the same applies to $\dualkarcher$ and $\dualjumps$.
This is addressed in for each variable separately in the following subsections.

In each iteration of ADMM, we subsequently minimize the augmented Lagrangian \eqref{eq:ltv:model-problem:Augmented-Lagrangian} with respect to one of the primal variables while fixing the others.
We choose $\bY, \bX, \assign, \bm$ as the order of updates in order to best exploit the problem structure.
Specifically, the $\bY$ and $\bX$ problems turn out to be independent of each other, and likewise the $\assign$ and $\bm$ problems are independent of each other as well.
We wish to point out that we did not encounter significant differences in the convergence behavior of the algorithm when changing the order of the updates.

We now address each of the four primal problems individually.
The index $\sequence{\cdot}{k}$ denotes the iteration number.
The overall ADMM algorithm is described in \cref{subsubsection:ltv-problem:ADMM-algorithm}.

\subsubsection{The \texorpdfstring{$\bY$}{Y}-Subproblem}
\label{subsubsection:ltv-problem:Y-subproblem}

The minimization of \eqref{eq:ltv:model-problem:Augmented-Lagrangian} with respect to~$\bY$ decouples into independent smooth problems, one on each triangle~$\triangle$.
Omitting the terms that do not depend on~$\bY$ as well as the factor $\frac{\rho}{2} \abs{T}$ common to all remaining terms, we obtain
\begin{equation}
	\label{eq:ltv-problem:Y-update}
	\underset{\bY_\triangle \in \paren(){\tangent{\bm_\triangle}[\Sphere]}^L}{\text{Minimize}}
	\quad
	\abs[Big]{
		\sum_{\ell=1}^L
		\sequence{\doubleindex \assign}{k} \doubleindex{\bY}
		+
		\sequence{\dualkarcher_\triangle}{k}
	}_2^2
	+
	\sum_{\ell=1}^L
	\abs[big]{
		\exponential{\sequence{\bm_\triangle}{k}}(\doubleindex{\bY})
		-
		\gl
		+
		\sequence{\doubleindex{\dualextra}}{k}
	}_2^2
	.
\end{equation}
We solve these problems simultaneously by applying a (Euclidean) gradient descent scheme with an Armijo backtracking strategy in the space $\DGat{\triangles}{\tangentBundle[\Sphere]}{\bm}$.
In fact, we check a stopping criterion per triangle and drop the converged triangles from subsequent gradient steps.
As stopping criterion we use $\abs{\nabla_{\bY_{\triangle}} \cL_\rho} \le \max \set[big]{10^{-8}, \; 10^{-0.0025 \, k}}$, dependent on the iteration number~$k$ of the outer ADMM loop.
The overall solution of the $\bY$-subproblem obtained in this way is denoted by $\sequence{\bY_\triangle}{k+1}$.

\subsubsection{The \texorpdfstring{$\bX$}{X}-Subproblem}
\label{subsubsection:ltv-problem:X-subproblem}

The minimization of \eqref{eq:ltv:model-problem:Augmented-Lagrangian} with respect to~$\bX$ also decouples into independent problems, one on each edge~$\edge$.
Omitting the terms that do not depend on~$\bX$ as well as the factor $\abs{E}$ common to all remaining terms, we obtain
\begin{equation}
  \underset{\bX_\edge \in \tangent{\sequence{\eplus{\bm}}{k}}[\Sphere]}{\text{Minimize}}
	\quad
	\TVweight \, \abs{\bX_\edge}_2
	+
	\frac{\rho}{2}
	\abs[big]{
    \logarithm{\sequence{\eplus{\bm}}{k}}(\sequence{\eminus{\bm}}{k})
		- \bX_\edge
    + \sequence{\dualjumps_\edge}{k}
	}_2^2
  .
\end{equation}
This problem is well known to have a closed-form solution in terms of a vector-valued soft-thresholding operation; see for instance \cite{GoldsteinOsher:2009:1}.
We obtain
\begin{equation}
	\label{eq:ltv-problem:X-update}
	\sequence{\bX_\edge}{k+1}
	=
	\paren[bigg](){
		1
		- \frac
		{\TVweight / \rho}
		{
			\max \paren[big]\{\}
			{
				\TVweight/ \rho
				, \;
				\abs[big]{
					\logarithm[big]{\sequence{\eplus{\bm}}{k}}(\sequence{\eminus{\bm}}{k})
					+
					\sequence{\dualjumps_\edge}{k}
				}_2
			}
		}
	}
	\paren[Big](){
		\logarithm[big]{\sequence{\eplus{\bm}}{k}}(\sequence{\eminus{\bm}}{k})
		+
		\sequence{\dualjumps_\edge}{k}
	}
	.
\end{equation}
The result $\sequence{\bX_\edge}{k+1}$ is naturally an element of the tangent space $\tangent{\eplus{\bm}}[\!\!\Sphere]$, because both $\logarithm[big]{\sequence{\eplus{\bm}}{k}}(\sequence{\eminus{\bm}}{k})$ and $\sequence{\dualjumps_\edge}{k}$ are.

\subsubsection{The \texorpdfstring{$\assign$}{phi}-Subproblem}
\label{subsubsection:ltv-problem:phi-subproblem}

Concerning $\assign$, the minimization of \eqref{eq:ltv:model-problem:Augmented-Lagrangian} decouples into independent quadratic programming problems (QPs), one on each triangle~$\triangle$.
Omitting the terms that do not depend on~$\assign$ as well as the factor $\abs{T}$ common to all remaining terms, we obtain
\begin{equation}
	\label{eq:ltv-problem:phi-update}
	\underset{\assign_\triangle \in \Simplex}{\text{Minimize}}
	\quad
	\assign_\triangle^\transp \similarity_\triangle
	+
	\frac{\rho}{2}
	\abs[Big]{
		\sum_{\ell=1}^L
		\doubleindex{\assign} \sequence{\doubleindex{\bY}}{k+1}
		+
		\sequence{\dualkarcher_\triangle}{k}
	}_2^2
	.
\end{equation}
This is a standard convex QP with constraints $\assign^\transp \bone = 0$ and $\assign \ge \bnull$.
We initially solved these problems using the QP solver \osqp from \cite{StellatoBanjacGoulartBemporadBoyd:2020:1}.
However, we found that a specialized solver exploiting the geometry of the probability simplex can be more efficient, particularly in light of the fact that we need to solve one problem of type \eqref{eq:ltv-problem:phi-update} for each triangle and each iteration of the ADMM algorithm, which may easily amount to millions of QPs solved.
One such approach can be found in \cite{LiMcKenzieYin:2021:1}, where the authors propose the substitution $\doubleindex{\bz}^2 = \doubleindex{\assign}$ and solve the problem with $z \in \Sphere_L = \setDef{\bx \in \R^L}{\abs{\bx}_2 = 1}$ using the Riemannian manifold structure of the sphere.
However, we propose the two-stage procedure described below, which exploits the fact that in later ADMM iterations the activity structure of the inequalities $\sequence{\assign}{k+1} \ge 0$ is unchanged from the previous iterate $\sequence{\assign}{k}$.

We proceed in a two-stage procedure to solve \eqref{eq:ltv-problem:phi-update}.
In the first stage, we apply a Riemannian gradient descent scheme with an Armijo backtracking strategy on the open probability simplex $\opensimplex$.
This makes use of the exponential map \eqref{eq:simplex:exponential-map} and of the Riemannian gradient \eqref{eq:simplex:Riemannian-gradient} of the objective in \eqref{eq:ltv-problem:phi-update}.
The gradient scheme stops when the iterates approach the boundary, \ie, when a component $\labelindex{\assign}$ becomes small, or when the norm of the Riemannian gradient falls below a threshold.

The approximate solution obtained in this way serves to determine the inactive and active inequalities, \ie, the subsets $\cI, \cA \subseteq \set{1, \ldots, L}$, where $\doubleindex{\assign} = 0$ on~$\cA$.
With the active set $\cA$ and its complement$~\cI$ at hand, the KKT system for \eqref{eq:ltv-problem:phi-update} becomes a linear system of equations,
\begin{equation}
	\label{eq:ltv-problem:phi-update:KKT-conditions-active-set}
	\begin{aligned}
		\bnull
		&
		=
		\rho \, \bY_\triangle \bY_\triangle^\transp \assign_\triangle
		+
		\rho \, \bY_\triangle \dualkarcher_\triangle
		+
		\similarity_\triangle
		-
		\balpha_\triangle
		-
		\gamma_\triangle \bone
		,
		\\
		1
		&
		=
		\bone^\transp \assign_\triangle
		,
		\\
		0
		&
		=
		\doubleindex{\assign}
		\quad
		\text{for }
		\ell \in \cA
		,
		\\
		0
		&
		=
		\doubleindex{\balpha}
		\quad
		\text{for }
		\ell \in \cI
		.
	\end{aligned}
\end{equation}
The variables $\balpha_\triangle \in \R^L$ and $\gamma_\triangle \in \R$ are the Lagrange multipliers associated with the constraints $\doubleindex{\assign} = 0$ and $\bone^\transp \assign_\triangle = 1$, respectively.
If a solution $(\assign_\triangle, \balpha_\triangle, \gamma_\triangle)$ to the KKT system \eqref{eq:ltv-problem:phi-update:KKT-conditions-active-set} exists and fulfills $\assign_\triangle \ge \bnull$ (primal feasibility) and $\balpha_\triangle \ge \bnull$ (dual feasibility), then $\assign_\triangle$ is in fact a solution to the original problem \eqref{eq:ltv-problem:phi-update}.
We then assign it to $\sequence{\assign_\triangle}{k+1}$.
Otherwise, the estimate of the active set was incorrect and we use the result of the Riemannian gradient descent stage for $\sequence{\assign_\triangle}{k+1}$ as a fallback.
The entire procedure is shown in \cref{algorithm:ltv-problem:assignment-subproblem}.

\begin{algorithm}[htb]
	\caption{Solution of the simplex-constrained QP \eqref{eq:ltv-problem:phi-update}, the $\assign$-subproblem}
	\label{algorithm:ltv-problem:assignment-subproblem}
	\begin{algorithmic}[1]
		\Require
		$\similarity_\triangle \in \R^L$,
		$\bY_\triangle \in(\tangent{\bm_\triangle}[\Sphere])^L$,
		$\dualkarcher_\triangle \in \tangent{\bm_\triangle}[\Sphere]$,
		\Require
		$\sequence{\assign_\triangle}{0} \in \Simplex$,
		$\varepsilon > 0$,
		$\text{TOL}_1, \text{TOL}_2, \text{TOL}_3 > 0$
		\Ensure
		approximate solution of \eqref{eq:ltv-problem:phi-update}
		\State
		Set
		$j \gets 0$
		\State
		Set
		$\sequence{\assign_\triangle}{0} \gets \frac{1}{1 + \varepsilon L} \paren[big](){\sequence{\assign_\triangle}{0} + \varepsilon \bone}$
		\Comment{recentralize $\sequence{\assign_\triangle}{0}$ to the open simplex $\opensimplex$}
		\While{
			$\abs[big]{\gradsimplex f_\triangle \paren[big](){\sequence{\assign_\triangle}{j}}}> \text{TOL}_1 $
			and
			$\min \setDef[big]{\sequence{\doubleindex{\assign}}{j}}{1 \le \ell \le L} > \text{TOL}_2$
		}
		\State
		Set
		$
		\sequence{\assign_\triangle}{j+1}
		\gets
		\exponential[big]{\sequence{\assign_\triangle}{j}}(- \sequence{\balpha_\triangle}{j+1} \gradsimplex f_\triangle \paren[big](){\sequence{\assign_\triangle}{j}})
		$
		\State
		Set
		$j \gets j+1$
		\EndWhile
		\Comment{end of stage~$1$}
		\State
		$\cA = \setDef[big]{1 \le \ell \le L}{\sequence{\doubleindex{\assign}}{j} > \text{TOL}_3}$
		\State
		Solve the KKT system \eqref{eq:ltv-problem:phi-update:KKT-conditions-active-set} for $(\assign_\triangle, \balpha_\triangle, \gamma_\triangle)$
		\If{$\balpha_\triangle, \assign_\triangle \ge \bnull$}
		\State
		\Return $\assign_\triangle$
		\Comment{solution of \eqref{eq:ltv-problem:phi-update}}
		\Else
		\State
		\Return $\sequence{\assign_\triangle}{j}$
		\Comment{return the result of stage~1 as fallback}
		\EndIf
	\end{algorithmic}
\end{algorithm}

\subsubsection{The \texorpdfstring{$\bm$}{m}-Subproblem}
\label{subsubsection:ltv-problem:m-subproblem}

It remains to discuss the update for the variable~$\bm$ representing the Riemannian center of mass \eqref{eq:sphere:Karcher-mean-problem}.
Omitting the terms that do not depend on~$\bm$ as well as the factor $\frac{\rho}{2}$ common to all remaining terms, we obtain
\begin{multline*}
	\underset{\bm \in \DG{\triangles}{\Sphere}}{\text{Minimize}}
	\quad
	\sum_{\triangle \in \triangles}
	\abs{\triangle}
	\sum_{\ell=1}^L
	\abs[big]{
		\exponential[big]{\bm_\triangle}(\sequence{\doubleindex{\bY}}{k+1})
		-
		\gl
		+
    \sequence{\doubleindex{\dualextra}}{k}
	}_2^2
	\\
	+
	\sum_{\edge \in \edges}
	\abs{\edge}
	\abs[big]{
    \logplusminus
    - \sequence{\bX_\edge}{k+1}
    + \sequence{\dualjumps_\edge}{k}
	}_2^2
	.
\end{multline*}
Note, however, that a change of the variable $\bm$ always requires a simultaneous update of the tangent vectors $\bY, \bX, \dualkarcher, \dualjumps$ to the tangent space of~$\Sphere$ at~$\bm$.
We therefore consider these variables to be dependent on~$\bm$ and replace them by their parallelly transported counterparts, leading to
\begin{multline}
	\label{eq:ltv-problem:m-update}
	\underset{\bm \in \DG{\triangles}{\Sphere}}{\text{Minimize}}
	\quad
	\sum_{\triangle \in \triangles}
	\abs{\triangle}
	\sum_{\ell=1}^L
	\abs[big]{%
		\exponential[big]{\bm_\triangle}(\parallelTransport[big]{\sequence{\bm_\triangle}{k}}{\bm_\triangle}(\sequence{\doubleindex{\bY}}{k+1}))
		-
		\gl
		+
    \sequence{\doubleindex{\dualextra}}{k}
	}_2^2
	\\
	+
	\sum_{\edge \in \edges}
	\abs{\edge}
	\abs[big]{
    \logplusminus
    +
		\parallelTransport[big]{\sequence{\eplus{\bm}}{k}}{\bm_\triangle}
		(- \sequence{\bX_\edge}{k+1} + \sequence{\dualjumps_\edge}{k})
	}_2^2
	.
\end{multline}
The explicit representation of the parallel transport was given in \eqref{eq:parallel-transport}.

Notice that problem \eqref{eq:ltv-problem:m-update} does not decouple over mesh entities since it contains both a sum over triangles and a sum over edges.
It is therefore the most expensive subproblem in the ADMM loop.
In the remainder of this subsection, we describe two approaches to approximately solve \eqref{eq:ltv-problem:m-update}.
The first method is a straightforward Riemannian gradient descent scheme with an Armijo backtracking strategy.
The second approach is a novel manifold Newton scheme recently devised in \cite{WeiglSchiela:2024:1} to find zeros in vector bundles and proven to converge superlinearly.
As demonstrated in the experiments in \cref{section:numerical-examples}, this can lead to a significant speedup compared to the gradient descent approach.

We denote the approximate solution of the $\bm$-subproblem \eqref{eq:ltv-problem:m-update} obtained in either way by $\sequence{\bm}{k+1}$.
We also update the tangent vectors $\sequence{\bY}{k+1}, \sequence{\bX}{k+1}, \sequence{\dualjumps}{k}$ by their parallely transported counterparts in their respective tangent spaces pertaining to the new iterate~$\sequence{\bm}{k+1}$.

\subsubsection*{Riemannian Gradient Descent for \eqref{eq:ltv-problem:m-update}}

For simplicity, we denote the objective in \eqref{eq:ltv-problem:m-update} by~$\mobjective$.
The gradient descent scheme is straightforward.
It uses the negative Riemannian gradient~$\delta \bm = -\riemanniangrad{\mobjective}\in \DGat{\triangles}{\tangentBundle[\Sphere]}{\bm}$ as a descent direction~$\delta \bm$.
The next iterate is then computed via the exponential map as $\exponential{\bm}(\stepsize \, \delta \bm)$, where the step size~$\stepsize > 0$ is determined via an Armijo backtracking strategy.
Given line-search parameters $\sigma \in \interval(){0}{1}$ and $\beta \in \interval(){0}{1}$, the step size is chosen as follows:
Starting with an initial stepsize $\stepsize$, we iteratively reduce $\stepsize$ by the factor~$\beta$ until the Armijo condition
\begin{equation}
  \label{eq:ltv:m-update:armijo-condition-gradient}
  \mobjective(\exponential{\bm}(\stepsize \, \delta \bm))
  \le
  \mobjective(\bm)
  +
  \sigma \, \stepsize \,
  \riemannian[big]{\riemanniangrad{\mobjective}}{\delta \bm}
\end{equation}
is fulfilled.
The procedure is summarized in~\cref{algorithm:ltv-problem:m-update:gradient}.
\begin{algorithm}[htb]
  \caption{Riemannian gradient descent for the $\bm$-subproblem \eqref{eq:ltv-problem:m-update}}
	\label{algorithm:ltv-problem:m-update:gradient}
	\begin{algorithmic}[1]
		\Require
		labels $\labels_1, \ldots, \labels_L \in \Sphere$,
    \Require
    Armijo parameters $\sigma \in (0,1)$, $\beta \in (0,1)$,
    \Require
    $\bm \in \DG{\triangles}{\Sphere}$
		\Require
		$\bY \in \DGat{\triangles}{\tangentBundle[\Sphere]}{\bm}^L$,
		$\bX \in \DGat{\edges}{\tangentBundle[\Sphere]}{\bm}$
    \Require
    $\dualextra \in \DG{\triangles}{\R^{L\times 3}}$,
		$\dualjumps \in \DGat{\edges}{\tangentBundle[\Sphere]}{\bm}$
		\Ensure
    approximate solution of \eqref{eq:ltv-problem:m-update}
    \State
    $\stepsize \gets 1$
		\label{step:ltv-problem:m-update:gradient:initialstepsize}
		\While{not converged}
		\State
    $\delta \bm \gets - \riemanniangrad{\mobjective}(\bm)$
		\label{step:ltv-problem:m-update:gradient:direction}
    \State
    $\stepsize \gets \min \set{1,2\stepsize}$
    \While{\eqref{eq:ltv:m-update:armijo-condition-gradient} is not fulfilled}
      \State
      $\stepsize \gets \beta \, \stepsize$
      \EndWhile
    \State
    $\bm \gets
    \exponential{\bm}
    \paren[auto](){\stepsize \, \delta \bm}$
		\EndWhile
    \State
    \Return $\bm$
	\end{algorithmic}
\end{algorithm}

Note that alternatively to starting with the initial stepsize $\stepsize=1$ in \cref{step:ltv-problem:m-update:gradient:initialstepsize},
we can also use the previously accepted stepsize $\stepsize$ in the previous ADMM iteration $k-1$ as an initial guess for the stepsize~$\stepsize$.
Either way, the iteration is stopped when a maximum of $10$~gradient steps is reached, or when the condition $\abs{\nabla_{\bm} \mobjective} \le \max \set[big]{10^{-8}, \; 10^{-0.0025 \, k}}$ is fulfilled.

\subsubsection*{Manifold Newton Method for \eqref{eq:ltv-problem:m-update}}

As an alternative to the Riemannian gradient descent scheme, we present a novel manifold Newton method.
It is based on \cite{WeiglSchiela:2024:1} devised to finding zeros in vector bundles.
We apply it to the first-order optimality condition of~\eqref{eq:ltv-problem:m-update} \wrt the optimization variable $\bm \in \DG{\triangles}{\Sphere}$, \ie, we seek $\bm^* \in \DG{\triangles}{\Sphere}$ satisfying
\begin{equation}
  \label{eq:ltv-problem:m-update:optimality-condition}
  \riemannianderivative{\mobjective}
  \paren[auto](){\bm^*}
  =
  \bnull
  \in
  \DGat{\triangles}{\cotangentBundle[\Sphere]}{\bm^*}
  .
\end{equation}
Notice that, due to the manifold structure of problem~\eqref{eq:ltv-problem:m-update}, the space~$\DGat{\triangles}{\cotangentBundle[\Sphere]}{\bm^*}$ depends on the unknown minimizer~$\bm^*$.

To formulate Newton's method, we need to derive a suitable Newton equation similiar to the classical Newton equation $\mobjectiveprimeprime(\bm) \, \delta \bm = - \mobjectiveprime(\bm)$ on linear spaces to compute Newton directions $\delta \bm \in \DGat{\triangles}{\tangentBundle}{\bm}$.
However, the second derivative~$\mobjectiveprimeprime(\bm)$ at~$\bm \in \DG{\triangles}{\Sphere}$ is a linear mapping $\mobjectiveprimeprime(\bm) \colon \DGat{\triangles}{\tangentBundle[\Sphere]}{\bm} \to \tangentSpace{\mobjectiveprime(\bm)}[\paren[big](){\DGat{\triangles}{\cotangentBundle[\Sphere]}{\bm}}]$.
On the other hand, the right-hand side in Newton's equation $-\mobjectiveprime(\bm) \in \DGat{\triangles}{\cotangentBundle[\Sphere]}{\bm}$ is not an element of the same space.
This mismatch is resolved by choosing a dual connection $\dualConnection{\ell} \colon \tangentSpace{\ell}[\paren(){\DGat{\triangles}{\cotangentBundle[\Sphere]}{\bm}}] \to \DGat{\triangles}{\cotangentBundle[\Sphere]}{\bm}$ for $\ell \in \DGat{\triangles}{\cotangentBundle[\Sphere]}{\bm}$.
That said, we can formulate the manifold Newton equation for \eqref{eq:ltv-problem:m-update:optimality-condition} as
\begin{equation}
  \label{eq:ltv-problem:m-update:newton-equation}
  \dualConnection{\mobjectiveprime(\bm)}
  \circ
  \mobjectiveprimeprime(\bm)
	\,
  \delta \bm
  =
  - \mobjectiveprime(\bm)
  \in
  \DGat{\triangles}{\cotangentBundle[\Sphere]}{\bm}
  .
\end{equation}

We now describe how to define a dual connection~$\dualconnection$ and evaluate the first and second derivatives $\mobjectiveprime, \mobjectiveprimeprime$ of the objective function~$\mobjective$ of \eqref{eq:ltv-problem:m-update} so that we can assemble and solve~\eqref{eq:ltv-problem:m-update:newton-equation}.
We closely follow the approach proposed in \cite[Section~5]{WeiglBergmannSchiela:2025:1}, \ie we use the Euclidean orthogonal projection from the Riemannian embedding of~$\Sphere$ into $\R^3$ to define a covector back-transport~$\vectorTransport{\cdot}{\bm}[*]$.
We then obtain a dual connection~$\dualconnection$ by differentiating this back-transport.

First, using the pointwise embedding $\iota$ of $\tangentBundle[\Sphere]$ into $\R^3$, we can define a pointwise vector transport and corresponding covector back-transport on $\cotangentBundle[\Sphere]$ using orthogonal projections $\orthogonalprojection(\bm) \colon \R^3 \to \tangentSpace{\bm}[\Sphere]$ for $\bm \in \Sphere$.
Here, the orthogonal projection is given by $\orthogonalprojection(\bm) \coloneqq \bI_3 - \bm \, \bm^\transp$, where $\bI_3 \in \R^{3\times 3}$ denotes the identity matrix.
We denote the space of linear maps between vector spaces by~$\linearmaps{\cdot}{\cdot}$.
Setting
\begin{equation*}
	\vectorTransport{\bm}{\eta}
	\coloneqq
	\orthogonalprojection(\eta) \, \iota(\bm)
	\in
	\linearmaps{\tangentSpace{\bm}[\Sphere]}{\tangentSpace{\eta}[\Sphere]}
\end{equation*}
for $\eta \in \Sphere$, we obtain a vector transport on the tangent bundle~$\tangentBundle[\Sphere]$.
With this, we can easily construct a vector transport $\vectorTransport{\bm}{\widetilde{\bm}} \colon \DGat{\triangles}{\tangentBundle[\Sphere]}{\bm} \to \DGat{\triangles}{\tangentBundle[\Sphere]}{\widetilde{\bm}}$ in a triangle-wise fashion by setting
\begin{equation*}
  \paren(){\vectorTransport{\bm}{\widetilde{\bm}}(\delta \bm)}_\triangle
  \coloneqq
  \vectorTransport{\bm_\triangle}{\widetilde{\bm}_\triangle}(\delta \bm_\triangle)
  =
  \orthogonalprojection(\widetilde{\bm}_\triangle) \, \iota(\bm_\triangle) \, \delta \bm_\triangle
  \text{ for all } \triangle \in \triangles
  ,
\end{equation*}
where $\delta \bm \in \DGat{\triangles}{\tangentBundle[\Sphere]}{\bm}$, \ie $\delta \bm_\triangle \in \tangentSpace{\bm_\triangle}[\Sphere]$.

The corresponding covector back-transport~$\vectorTransport{\cdot}{\bm}[*]$ on the cotangent bundle~$\DGat{\triangles}{\cotangentBundle[\Sphere]}{\bm}$ is now given by the dual map
\begin{equation*}
	\dual{\vectorTransport{\widetilde{\bm}}{\bm}(\ell)[*]}{\bv}
  =
  \dual{\ell}{\vectorTransport{\bm}{\widetilde{\bm}}(\bv)}
  \quad
  \text{for all }
  \ell \in \DGat{\triangles}{\cotangentBundle[\Sphere]}{\widetilde{\bm}}
	\text{ and }
  \bv \in \DGat{\triangles}{\tangentBundle[\Sphere]}{\bm}
	.
\end{equation*}

We now pass to the dual connection $\dualConnection{\ell}$ at $\ell \in \DGat{\triangles}{\cotangentBundle[\Sphere]}{\bm}$, obtained by differentiating the covector back-transport $\vectorTransport{\cdot}{\bm}(\ell)[*]$; see \cite[Section~5]{WeiglBergmannSchiela:2025:1}.
Specifically, we only need to know $\dualConnection{\mobjectiveprime(\bm)} \circ \mobjectiveprimeprime(\bm)$.
To this end, we apply the techniques from \cite[Proposition~4.3]{WeiglBergmannSchiela:2025:1} in a pointwise fashion.
Interpreting the objective $\mobjective$ defined in~\eqref{eq:ltv-problem:m-update} as a function $\mobjective \colon \DG{\triangles}{\R^3} \to \R$, we denote its derivatives by $\euclideanderivative{\mobjective}$ and $\euclideansecondderivative{\mobjective}$.
Then, the left-hand side in~\eqref{eq:ltv-problem:m-update:newton-equation} can be evaluated as
\begin{multline*}
	\dualConnection{\mobjectiveprime(\bm)}
	\mobjectiveprimeprime(\bm)
	\,
  \delta \bm
	\\
  =
  \euclideansecondderivative{\mobjective}(\bm) \, \delta \bm
  +
  \euclideanderivative{\mobjective}(\bm)
  \paren[auto](){\iota \circ \vectorTransport{\bm}{\bm}}'(\bm)
	\,
  \iota(\bm)
	\,
  \delta \bm
  \in
  \DGat{\triangles}{\cotangentBundle[\Sphere]}{\bm}
	,
\end{multline*}
which corresponds to \cite[Eq.~(22)]{WeiglBergmannSchiela:2025:1}.

It remains to address the Euclidean derivatives $\euclideanderivative{\mobjective}, \euclideansecondderivative{\mobjective}$ of the extended objective function~$\mobjective$ and $\paren(){\vectorTransport{\bm}{\bm}}'$.
These can be computed using automatic differentiation tools or using simple calculus.
However, the derivatives of the logarithmic terms $\logplusminus$ in~\eqref{eq:ltv-problem:m-update} require special treatment in case $\eplus{\bm} \approx \eminus{\bm}$, which is achieved by using the robust formulas \eqref{eq:log-derivative:h-definition}--\eqref{eq:log-derivative:limits}.

We have now defined all components required to assemble the Newton equation~\eqref{eq:ltv-problem:m-update:newton-equation}.
Although we will be representing the tangent and cotangent vectors in the embedding~$\Sphere \subseteq \R^3$, we need to keep in mind the fact that \eqref{eq:ltv-problem:m-update:newton-equation} is posed on the cotangent space~$\DGat{\triangles}{\cotangentBundle[\Sphere]}{\bm}$.
Thus, we only require the Newton direction~$\delta \bm \in \DGat{\triangles}{\tangentBundle[\Sphere]}{\bm}$ to fulfill~\eqref{eq:ltv-problem:m-update:newton-equation} \wrt all tangent vectors~$\delta \widetilde{\bm} \in \DGat{\triangles}{\tangentBundle[\Sphere]}{\bm}$, as opposed to all vectors in $\DG{\triangles}{\R^3}$.
Similar to \cite[Section~6]{WeiglBergmannSchiela:2025:1}, we address this by constructing a Euclidean-orthonormal basis~$\cB$ of the embedded tangent space~$\DGat{\triangles}{\tangentBundle[\Sphere]}{\bm}$ and $\cB^*$ of the cotangent space~$\DGat{\triangles}{\cotangentBundle[\Sphere]}{\bm}$.
Then, the linear system~\eqref{eq:ltv-problem:m-update:newton-equation} is represented with respect to the chosen bases~$\cB,\cB^*$.

We solve \eqref{eq:ltv-problem:m-update:newton-equation} using the \petsc implementation of the CG method.
We note that the Newton matrix in \eqref{eq:ltv-problem:m-update:newton-equation} is symmetric but not necessarily positive definite; see \cite[Chapter~XIII, §1]{Lang:1999:1}.
Therefore, the (truncated) CG iteration stops in case a direction of negative curvature is encountered, as described in \cite[Chapter~7.1]{NocedalWright:2006:1}.
We also exploit the fact that we do not need to solve \eqref{eq:ltv-problem:m-update:newton-equation} exactly in early ADMM iterations.
We therefore require a relative tolerance $\rtolnewtoncg = \min \set{0.1, \, \norm{\mobjectiveprime(\bm)}_{\DGat{\triangles}{\cotangentBundle[\Sphere]}{\bm}}^{1/2}}$
to accept a solution~$\delta \bm$ to the linear system~\eqref{eq:ltv-problem:m-update:newton-equation},
with the norm~$\norm{\cdot}_{\DGat{\triangles}{\cotangentBundle[\Sphere]}{\bm}}$ induced by the $L^2$-inner product on the cotangent bundle~$\DGat{\triangles}{\cotangentBundle[\Sphere]}{\bm}$.

This completes the description of the manifold Newton method for the $\bm$-update~\eqref{eq:ltv-problem:m-update}.
The whole procedure is summarized in \cref{algorithm:ltv-problem:m-update:riemannian-newton}.
\begin{algorithm}[htb]
  \caption{Manifold Newton method for the $\bm$-subproblem \eqref{eq:ltv-problem:m-update}}
	\label{algorithm:ltv-problem:m-update:riemannian-newton}
	\begin{algorithmic}[1]
		\Require
		labels $\labels_1, \ldots, \labels_L \in \Sphere$,
    \Require
    $\bm \in \DG{\triangles}{\Sphere}$
		\Require
		$\bY \in \DGat{\triangles}{\tangentBundle[\Sphere]}{\bm}^L$,
		$\bX \in \DGat{\edges}{\tangentBundle[\Sphere]}{\bm}$
    \Require
    $\dualextra \in \DG{\triangles}{\R^{L\times 3}}$,
		$\dualjumps \in \DGat{\edges}{\tangentBundle[\Sphere]}{\bm}$
		\Ensure
    approximate solution of \eqref{eq:ltv-problem:m-update}
		\While{not converged}
    \State Use truncated CG to compute an inexact Newton direction~$\delta \bm$ from \eqref{eq:ltv-problem:m-update:newton-equation}
    \label{step:ltv-problem:m-update:riemannian-newton:linsolve}
    \State
    \label{step:ltv-problem:m-update:riemannian-newton:exp}
    $\bm \gets \exponential{\bm} \delta \bm$
		\EndWhile
    \State
    \Return $\bm$
	\end{algorithmic}
\end{algorithm}

We found that in the setting of the ADMM scheme, even a single Newton step per ADMM iteration is sufficient.

\subsubsection{Overall ADMM Scheme}
\label{subsubsection:ltv-problem:ADMM-algorithm}

The overall ADMM scheme for the solution of the label space total variation problem \eqref{eq:ltv:model-problem} in the form of \eqref{eq:ltv:model-problem:expanded} is summarized in \cref{algorithm:ltv-problem:ADMM}.
As was mentioned before, due to the independence of the $\bY$ and $\bX$ problems, as well as the $\assign$ and $\bm$ problems, \cref{step:ltv-problem:ADMM:Y-update,step:ltv-problem:ADMM:X-update} as well as \cref{step:ltv-problem:ADMM:phi-update,step:ltv-problem:ADMM:m-update} can be executed in parallel.

\begin{algorithm}[htb]
	\caption{ADMM for the label space total variation problem \eqref{eq:ltv:model-problem:expanded}.}
	\label{algorithm:ltv-problem:ADMM}
	\begin{algorithmic}[1]
		\Require
		labels $\labels_1, \ldots, \labels_L \in \Sphere$,
		similarity measure $\similarity \in \DG{\triangles}{\R^L}$
		\Require
		TV penalty parameter~$\TVweight>0$,
		augmentation parameter~$\rho>0$
		\Require
		$\sequence{\assign}{0} \in \DG{\triangles}{\Simplex}$,
		$\sequence{\bm}{0} \in \DG{\triangles}{\Sphere}$,
		\Require
		$\sequence{\bY}{0} \in \DGat{\triangles}{\tangentBundle[\Sphere]}{\sequence{\bm}{0}}^L$,
		$\sequence{\bX}{0} \in \DGat{\edges}{\tangentBundle[\Sphere]}{\sequence{\bm}{0}}$
		\Ensure
		approximate solution of \eqref{eq:ltv:model-problem:expanded}
		\While{not converged}
		\State Use a gradient descent scheme to find an approximate minimizer $\sequence{\bY}{k+1}$ of \eqref{eq:ltv-problem:Y-update}; see \cref{subsubsection:ltv-problem:Y-subproblem}.
		\label{step:ltv-problem:ADMM:Y-update}
		\State Set $\sequence{\bX}{k+1}$ using the soft-thresholding operation \eqref{eq:ltv-problem:X-update}; see \cref{subsubsection:ltv-problem:X-subproblem}
		\label{step:ltv-problem:ADMM:X-update}
		\State Set $\sequence{\assign}{k+1}$ to the solution of \eqref{eq:ltv-problem:phi-update}, using \cref{algorithm:ltv-problem:assignment-subproblem}; see \cref{subsubsection:ltv-problem:phi-subproblem}
		\label{step:ltv-problem:ADMM:phi-update}
    \State Use \cref{algorithm:ltv-problem:m-update:gradient} (Riemannian gradient descent) or \cref{algorithm:ltv-problem:m-update:riemannian-newton} (manifold Newton) to find an approximate minimizer~$\sequence{\bm}{k+1}$ of \eqref{eq:ltv-problem:m-update}; see \cref{subsubsection:ltv-problem:m-subproblem}.
		This also updates $\sequence{\bY}{k+1}, \sequence{\bX}{k+1}, \sequence{\dualjumps}{k}$ via parallel transport.
		\label{step:ltv-problem:ADMM:m-update}
		\Statex Update the Lagrange multipliers:
		\State
		$
		\sequence{\dualkarcher_\triangle}{k+1}
		\gets
		\sequence{\dualkarcher_\triangle}{k}
		+
		\sum_{\ell=1}^L
		\sequence{\doubleindex{\assign}}{k+1} \sequence{\doubleindex{\bY}}{k+1}
		\quad
		\text{for all }
		\triangle \in \triangles
		$
		\label{step:ltv-problem:ADMM:dualkarcher-update}
		\State
		$
		\sequence{\doubleindex{\dualextra}}{k+1}
		\gets
		\sequence{\doubleindex{\dualextra}}{k}
		+
		\exponential[big]{\sequence{\bm_\triangle}{k+1}}(\sequence{\doubleindex{\bY}}{k+1})
		- \gl
		\quad
		\text{for all }
		\triangle \in \triangles
		\text{and }
		\ell = 1, \ldots, L
		$
		\label{step:ltv-problem:ADMM:dualextra-update}
		\State
		$
		\sequence{\dualjumps_\edge}{k+1}
		\gets
		\sequence{\dualjumps_\edge}{k}
		+
		\logarithm[big]{\sequence{\eminus{\bm}}{k+1}}(\sequence{\eplus{\bm}}{k+1})
		- \sequence{\bX_\edge}{k+1}
		\quad
		\text{for all }
		\edge \in \edges
		$
		\label{step:ltv-problem:ADMM:dualjumps-update}
		\EndWhile
	\end{algorithmic}
\end{algorithm}

\section{Numerical Examples}
\label{section:numerical-examples}

In this section, we compare the solutions of the proposed segmentation problems \eqref{eq:atv:model-problem} and \eqref{eq:ltv:model-problem} for a number of label sets and two different surfaces.
For each example, we proceed as follows.
We begin with a mesh~$\mesh$ and add Gaussian noise to the vertex positions.
The vertex-dependent variance is chosen as $\sigma^2 = 0.04 \, e^2$ and it scales with the average length~$e$ of the edges adjacent to the respective vertex.
We then solve \eqref{eq:atv:model-problem} using \cref{algorithm:atv-problem:Chambolle-Pock-algorithm} and \eqref{eq:ltv:model-problem} using \cref{algorithm:ltv-problem:ADMM}.

Since the two variants of total variation penalty are not directly comparable by value, we choose the penalty parameter~$\TVweight$ independently for each model and example.
In each case, we experimentally determine the optimal value~$\TVweight^*$ in the following way.
We solve the assignment problem with $\TVweight = 0$ for the mesh without noise, which simply amounts to finding the index where $\doubleindex{\similarity}$ is minimal on each triangle~$\triangle$.
We then pick $\TVweight^*$ as the value of $\TVweight$ that gives the best result in terms of the minimal number of incorrectly labeled triangles, weighted by triangle area.
We also show the results for larger and smaller values of~$\TVweight$ to show the sensitivity to variations of this parameter.

Besides the percentage of correctly labeled triangles (weighted by triangle area), we also report the Rand index~$\RI$ \cite{HubertArabie:1985:1,Rand:1971:1} to evaluate the quality of the segmentation results.
The $\RI$ compares two given partitions $X = \set{X_1, \ldots, X_k}$ and $Y = \set{Y_1, \ldots, Y_r}$ of a finite set $S = \set{s_1, \ldots, s_n}$.
It evaluates to $\RI = \frac{a}{b}$, where $a$ is the number of pairs $\set{s_i, s_j}$ that are either in the same subset in both partitions $X$ and $Y$, or in different subsets in both partitions.
Moreover, $b = \binom{n}{2}$ is the total number of pairs.

The assignment function $\assign$ takes values in the simplex~$\Simplex$ but not necessarily equal to a vertex.
That is, some triangles will be ambiguously labeled, which needs to be resolved both for the purpose of visualization, for the procedure to select the optimal penalty parameter~$\TVweight^*$, as well as for the evaluation of both performance metrics.
To this end, we simply choose the label with the highest assignment value per triangle in case of the assignment space total variation problem \eqref{eq:atv:model-problem}.
For the label space total variation problem \eqref{eq:ltv:model-problem}, we choose the label~$\gl$ with the smallest geodesic distance to the assigned normal~$\mphi_\triangle$.
We found this to be a more natural choice in view of the mapping $\assign_\triangle \mapsto \mphi[\assign_\triangle]$ not being injective so that several assignments $\assign_\triangle$ yield the same Riemannian center of mass~$\mphi[\assign_\triangle]$.

The code for the numerical experiments is written in \python using the \fenics finite element library version~2019.1 \cite{AlnaesBlechtaHakeJohanssonKehletLoggRichardsonRingRognesWells:2015:1}.
It was run in parallel using 15~cores on an Intel Quad Xeon CPU@\SI{2.6}{\giga\hertz} compute server.
The parallelization is based on the built-in \mpi support provided by \fenics, which essentially addresses the subproblems in \cref{algorithm:atv-problem:Chambolle-Pock-algorithm,algorithm:ltv-problem:ADMM} by way of domain decomposition.
We do not exploit the possible parallel execution of the update steps of $\bY$ and $\bX$ and, respectively, $\assign$ and $\bm$, as mentioned in \cref{subsubsection:ltv-problem:ADMM-algorithm}.

\Cref{table:runtimes} shows the approximate run time for all experiments in this section, each for the respective optimal value of~$\TVweight^*$.
In general, the computation time depends on the size of the mesh, the number of labels~$L$, and the TV regularization parameter~$\TVweight$.
As expected, the more complex label space assignment TV model \eqref{eq:ltv:model-problem} requires a higher computational effort than the assignment TV model \eqref{eq:atv:model-problem}.
On the other hand, the new Newton method (\cref{algorithm:ltv-problem:m-update:riemannian-newton}) provides a significant speedup compared to the gradient descent approach (\cref{algorithm:ltv-problem:m-update:gradient}) for the $\bm$-update step, which is the main computational bottleneck in the ADMM scheme.

\begin{table}[htb]
	\crefname{algorithm}{Alg.}{Algs.}
	\Crefname{algorithm}{Alg.}{Algs.}
  \centering
  \begin{tabular}{l r r rr}
    \toprule
    &
    &
    \eqref{eq:atv:model-problem}
    &
		\multicolumn{2}{c}{\eqref{eq:ltv:model-problem}}
		\\
    \cmidrule(lr){3-3}
    \cmidrule(lr){4-5}
		\multicolumn{2}{r}{number of labels $L$}
    &
		\cref{algorithm:atv-problem:Chambolle-Pock-algorithm}
    &
    \crefns{algorithm:ltv-problem:ADMM,algorithm:ltv-problem:m-update:gradient}
    &
    \crefns{algorithm:ltv-problem:ADMM,algorithm:ltv-problem:m-update:riemannian-newton}
    \\
    \midrule
    unit sphere mesh
    &
		$22$
    &
    \SI{83}{s}
    &
    \SI{3400}{s}
    &
    \SI{1661}{s}
    \\
    fandisk mesh, nonuniform labels
    &
		$29$
    &
    \SI{272}{s}
    &
    \SI{5737}{s}
    &
    \SI{3451}{s}
    \\
    fandisk mesh, uniform labels
    &
		$50$
    &
    \SI{138}{s}
    &
    \SI{8664}{s}
    &
    \SI{4760}{s}
    \\
    \bottomrule
  \end{tabular}
	\caption{%
		Approximate run times to solve the \eqref{eq:atv:model-problem} and \eqref{eq:ltv:model-problem} problems for their respective optimal regularization parameters~$\TVweight^*$; see \cref{table:unit-sphere-mesh:atv-ltv,table:fandisk-mesh:nonuniform:atv-ltv,table:fandisk-mesh:uniform:atv-ltv}.
		For the \eqref{eq:ltv:model-problem} problem solved with the ADMM \cref{algorithm:ltv-problem:ADMM}, we report timings for both gradient descent (\cref{algorithm:ltv-problem:m-update:gradient}) as well as Newton's method (\cref{algorithm:ltv-problem:m-update:riemannian-newton}) applied to solve the $\bm$-subproblem.
	}
	\label{table:runtimes}
\end{table}

\subsection{Unit Sphere Mesh}
\label{subsection:numerical-examples:unit-sphere-mesh}

\begin{figure}[htb]
	\centering
  \includegraphics[width = 0.3\textwidth]{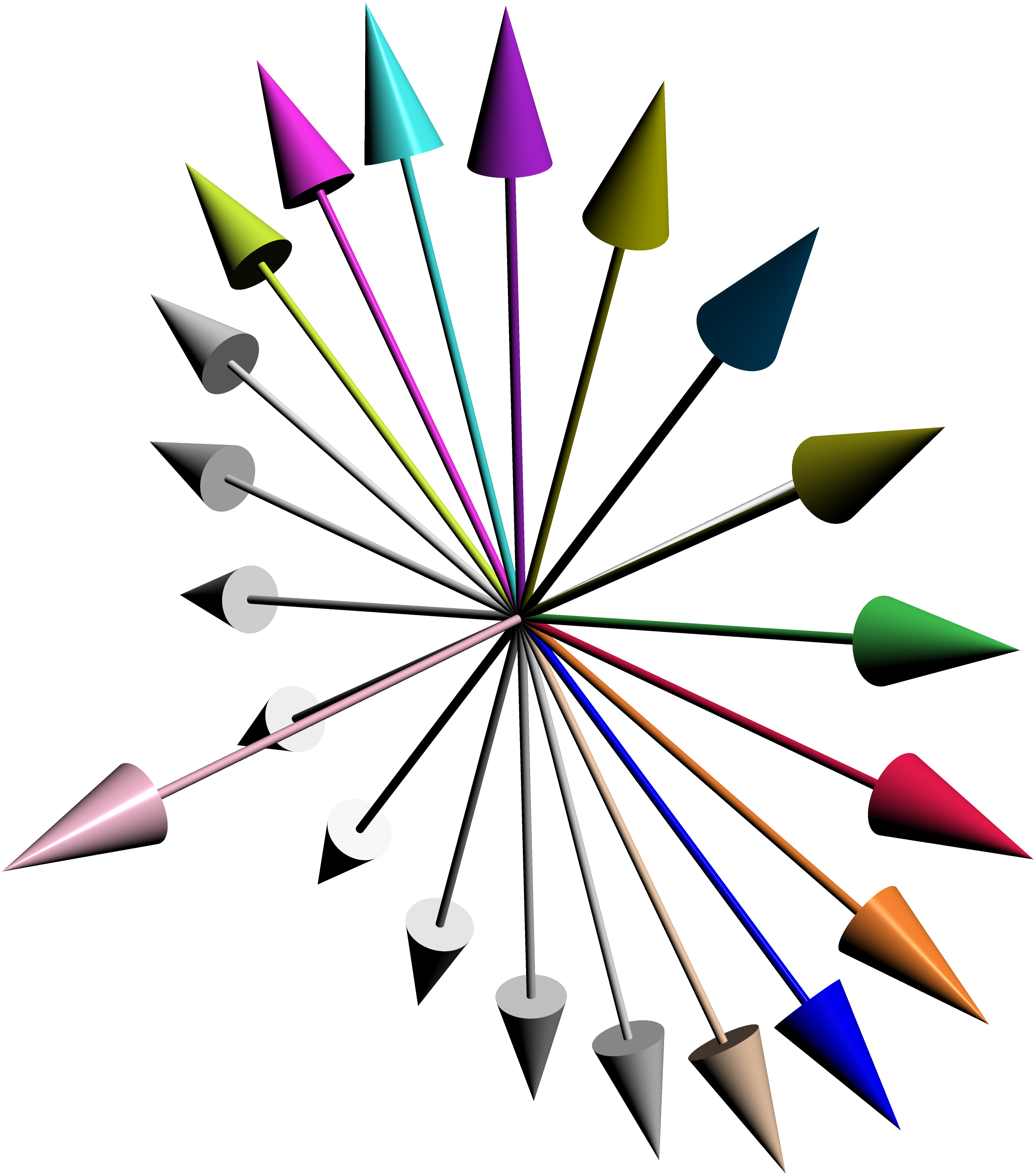}
  \caption{%
		Visualization of the label set \eqref{eq:unit-sphere-mesh:labels} for the sphere mesh example (\cref{subsection:numerical-examples:unit-sphere-mesh}).
		Labels that are not visible in the corresponding result plots~\cref{figure:unit-sphere-mesh:atv-ltv} are colored in white.
	}
	\label{figure:unit-sphere-mesh:labels}
\end{figure}

The first mesh we consider is a discretization of the unit sphere $\Sphere \subseteq \R^3$ into \num{4554}~triangles.
We choose the labels to form an equidistant partition around the equator, plus the poles, namely
\begin{equation}
	\label{eq:unit-sphere-mesh:labels}
	\labels_\ell
	\coloneqq
	\begin{pmatrix}
		\sin \paren[auto](){\ell \cdot \frac{2 \pi}{20}}
		\\
		\cos \paren[auto](){\ell \cdot \frac{2 \pi}{20}}
		\\
		0
	\end{pmatrix}
	\text{ for }
	\ell = 1, \ldots, 20
	,
	\;
	\labels_{21}
	=
	\begin{pmatrix}
		0
		\\
		0
		\\
		1
	\end{pmatrix}
	\text{and }
  \labels_{22}
  =
	- \labels_{21}
	.
\end{equation}
The labels are visualized in \cref{figure:unit-sphere-mesh:labels}.

\begin{figure}[htb]
	\centering
	\begin{subfigure}{0.31\textwidth}
		\begin{center}
			\makeatletter
			\ltx@ifclassloaded{amsart}{%
				\includegraphics[width = 0.9\textwidth]{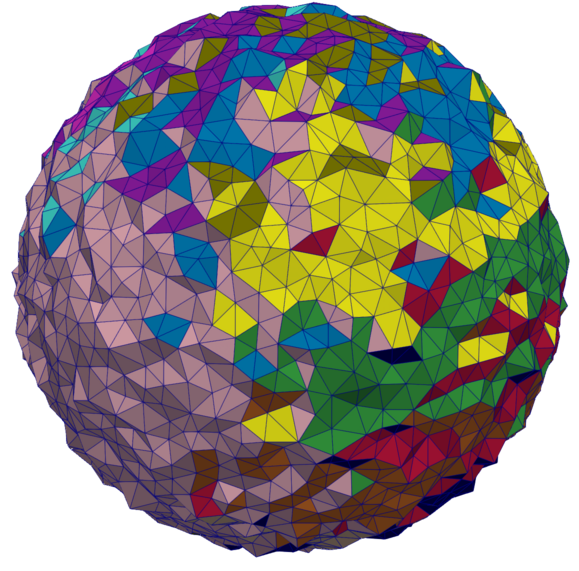}
			}{%
				\includegraphics[width = 0.9\textwidth]{pictures/sphere_disk20_assignmentTV_0_002.png}
			}
			\makeatother
		\end{center}
		\caption{$\TVweight = 0.002$}
		\label{figure:sphere-mesh:atv:small}
	\end{subfigure}
	\begin{subfigure}{0.31\textwidth}
		\begin{center}
			\makeatletter
			\ltx@ifclassloaded{amsart}{%
				\includegraphics[width = 0.9\textwidth]{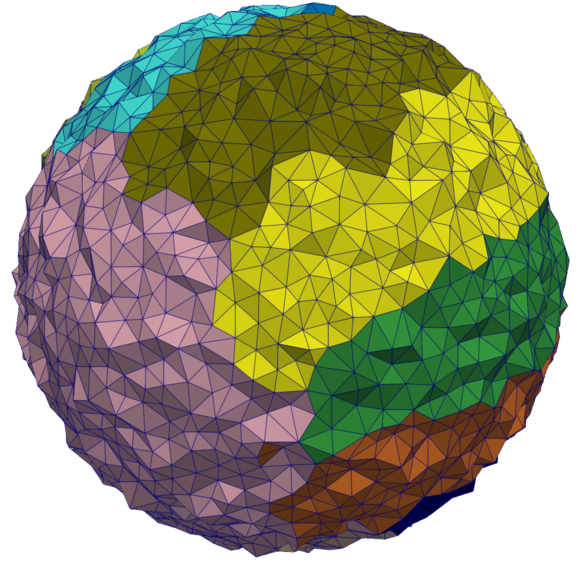}
			}{%
				\includegraphics[width = 0.9\textwidth]{pictures/sphere_disk20_assignmentTV_0_008.png}
			}
			\makeatother
		\end{center}
		\caption{$\TVweight = \TVweight^*_\atv = 0.008$}
		\label{figure:sphere-mesh:atv:optimal}
	\end{subfigure}
	\begin{subfigure}{0.31\textwidth}
		\begin{center}
			\makeatletter
			\ltx@ifclassloaded{amsart}{%
				\includegraphics[width = 0.9\textwidth]{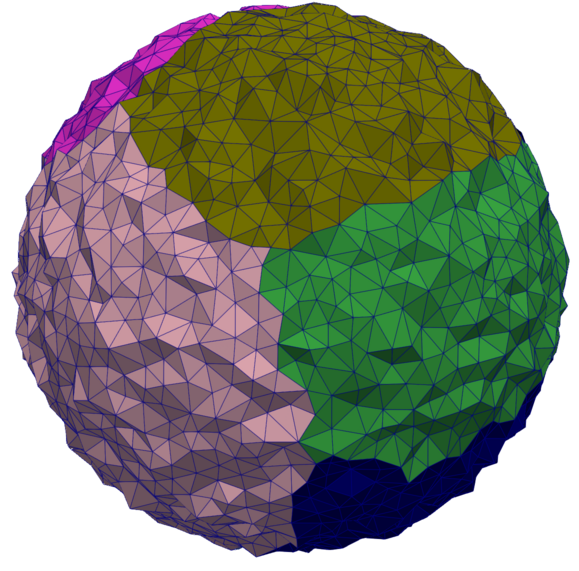}
			}{%
				\includegraphics[width = 0.9\textwidth]{pictures/sphere_disk20_assignmentTV_0_032.png}
			}
			\makeatother
		\end{center}
		\caption{$\TVweight = 0.032$}
		\label{figure:sphere-mesh:atv:large}
	\end{subfigure}
	\\[\baselineskip]
	\begin{subfigure}{0.31\textwidth}
		\begin{center}
			\makeatletter
			\ltx@ifclassloaded{amsart}{%
				\includegraphics[width = 0.9\textwidth]{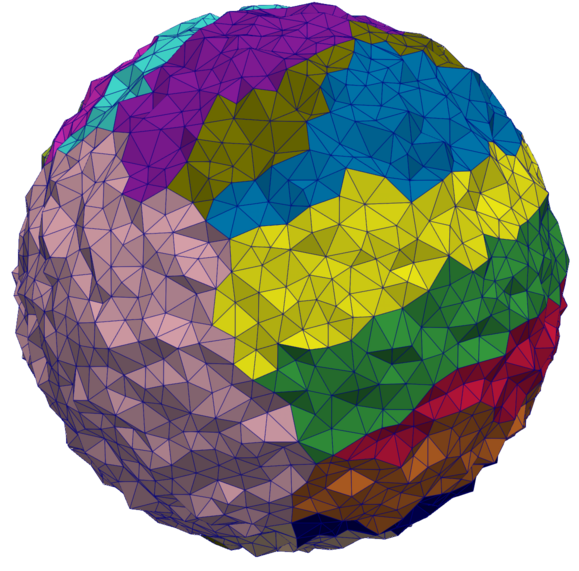}
			}{%
				\includegraphics[width = 0.9\textwidth]{pictures/sphere_disk20_labelTV_0_03125.png}
			}
			\makeatother
		\end{center}
		\caption{$\TVweight = 0.03125$}
		\label{figure:sphere-mesh:ltv:small}
	\end{subfigure}
	\begin{subfigure}{0.31\textwidth}
		\begin{center}
			\makeatletter
			\ltx@ifclassloaded{amsart}{%
				\includegraphics[width = 0.9\textwidth]{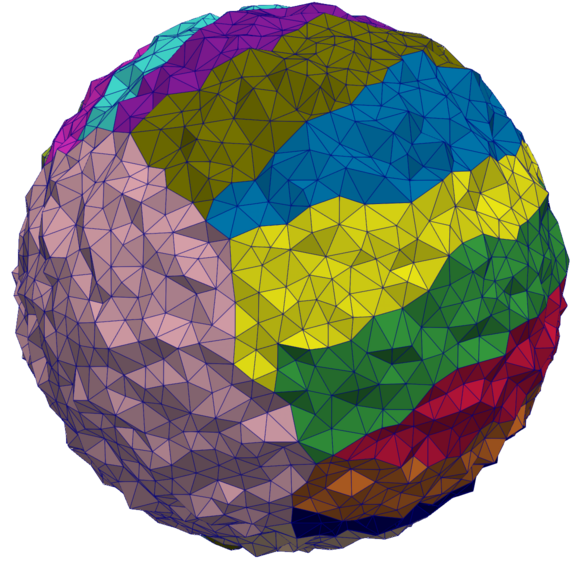}
			}{%
				\includegraphics[width = 0.9\textwidth]{pictures/sphere_disk20_labelTV_0_125.png}
			}
			\makeatother
		\end{center}
		\caption{$\TVweight = \TVweight^*_\ltv = 0.125$}
		\label{figure:sphere-mesh:ltv:optimal}
	\end{subfigure}
	\begin{subfigure}{0.31\textwidth}
		\begin{center}
			\makeatletter
			\ltx@ifclassloaded{amsart}{%
				\includegraphics[width = 0.9\textwidth]{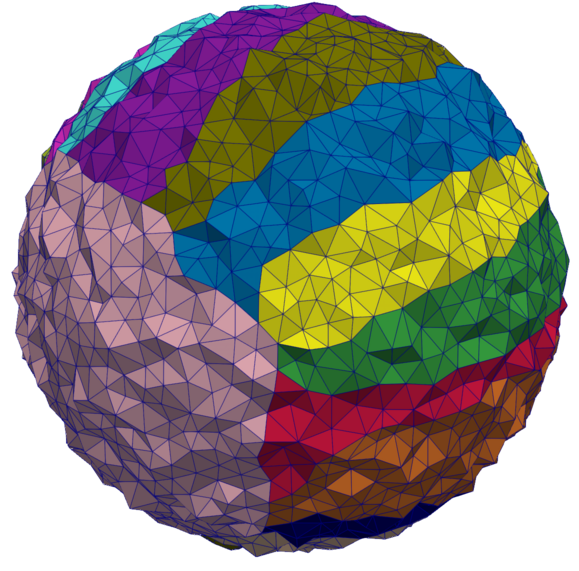}
			}{%
				\includegraphics[width = 0.9\textwidth]{pictures/sphere_disk20_labelTV_0_5.png}
			}
			\makeatother
		\end{center}
		\caption{$\TVweight = 0.5$}
		\label{figure:sphere-mesh:ltv:large}
	\end{subfigure}
	\caption{%
		Assignments for the noisy sphere mesh with different values of $\TVweight$ and the label set depicted in \cref{figure:unit-sphere-mesh:labels}.
		Triangles are colored according to their assigned label.
		One of the poles is shown in light pink.
		\Cref{figure:sphere-mesh:atv:small,figure:sphere-mesh:atv:optimal,figure:sphere-mesh:atv:large} show the results using \eqref{eq:atv:model-problem}, while \cref{figure:sphere-mesh:ltv:small,figure:sphere-mesh:ltv:optimal,figure:sphere-mesh:ltv:large} show the results using the new \eqref{eq:ltv:model-problem} model.
    Quantitative results for these examples are given in \cref{table:unit-sphere-mesh:atv-ltv}.
	}
	\label{figure:unit-sphere-mesh:atv-ltv}
\end{figure}

\Cref{figure:unit-sphere-mesh:atv-ltv} shows the resulting assignments for this setup.
For each of the two models, we choose $\beta \in \set{1/4, 1, 4} \cdot \TVweight^*$, where $\TVweight^*$ is the optimal parameter choice for the respective model.

For the optimal parameter~$\TVweight^*_\atv$, the solution of the \eqref{eq:atv:model-problem} model uses 21 out of the available $22$~labels.
However, 6 of the labels are used only on a very small number of triangles and are not visible in \cref{figure:unit-sphere-mesh:atv-ltv}.
Overall, \SI{69.4}{\percent} of all triangles are correctly labeled with a Rand index of $0.953$.
By contrast, the new \eqref{eq:ltv:model-problem} uses all $22$~labels evenly and achieves an improved correctness of \SI{85.2}{\percent} and a Rand index of $0.973$.
Further quantitative results are shown in \cref{table:unit-sphere-mesh:atv-ltv}.

\begin{table}[htb]
	\robustify\bfseries
	\centering
	\begin{tabular}{
			l
			l
      S[table-format = 1.5, table-model-setup = \bfseries,group-digits=false]
			S[table-format = 2]
			S[table-format = 2.1, table-model-setup = \bfseries]
			S[table-format = 1.3, table-model-setup = \bfseries]
		}
		\toprule
		&
		model
		&
		{TV weight $\TVweight$}
		&
		{labels used}
		&
		{correctly labeled~\unit{\percent}}
		&
		{RI}
		\\
		\midrule
		\cref{figure:sphere-mesh:atv:small}
		&
		\eqref{eq:atv:model-problem}
		&
		0.002
		&
		22
		&
		58.3
		&
		0.931
		\\
		\cref{figure:sphere-mesh:atv:optimal}
		&
		\eqref{eq:atv:model-problem}
		&
		\bfseries 0.008
		&
		21
		&
		\bfseries 69.4
		&
		\bfseries 0.953
		\\
		\cref{figure:sphere-mesh:atv:large}
		&
		\eqref{eq:atv:model-problem}
		&
		0.032
		&
		9
		&
		55.7
		&
		0.934
		\\
		\midrule
		\cref{figure:sphere-mesh:ltv:small}
		&
		\eqref{eq:ltv:model-problem}
		&
		0.03125
		&
		22
		&
		83.6
		&
		0.972
		\\
		\cref{figure:sphere-mesh:ltv:optimal}
		&
		\eqref{eq:ltv:model-problem}
		&
		\bfseries 0.125
		&
		22
		&
		\bfseries 85.2
		&
		\bfseries 0.973
		\\
		\cref{figure:sphere-mesh:ltv:large}
		&
		\eqref{eq:ltv:model-problem}
		&
		0.5
		&
		22
		&
		81.2
		&
		0.964
		\\
		\bottomrule
	\end{tabular}
	\caption{%
		Quantitative results for the noisy sphere mesh example (\cref{figure:unit-sphere-mesh:atv-ltv}) for different values of the regularization parameter~$\TVweight$ and the two models.
    The number of correctly labeled triangles and the Rand index are measured with respect to the ground-truth segmentation given by the original mesh without noise and without regularization ($\TVweight = 0$).
  }
	\label{table:unit-sphere-mesh:atv-ltv}
\end{table}

The sphere mesh together with our choice of labels illustrates that the label space total variation model \eqref{eq:ltv:model-problem} incurs no additional regularization penalties for \enquote{in-between labels}; see also \cref{example:comparison-of-the-regularizers} and the discussion there.
Therefore, it is able to use all available labels.

A second benefit of the new model is its robustness with respect to the choice of the regularization parameter.
In case of too small a regularization parameter (\cref{figure:sphere-mesh:atv:small}), we see that \eqref{eq:atv:model-problem} fails to produce segment the mesh in a meaningful way, in contrast to \eqref{eq:ltv:model-problem}.
For too large a regularization parameter (\cref{figure:sphere-mesh:atv:large}), the assignment space total variation model \eqref{eq:atv:model-problem} does produce meaningful segments, but only uses a subset of the of the available labels, while the label space total variation model \eqref{eq:ltv:model-problem} still uses all~22.
Again, we refer the reader to \cref{table:unit-sphere-mesh:atv-ltv} for details.

\subsection{Fandisk Mesh}
\label{subsection:numerical-examples:fandisk-mesh}

\begin{figure}[htb]
	\centering
  \begin{subfigure}{0.45\textwidth}
    \centering
    \includegraphics[width = 0.8\textwidth]{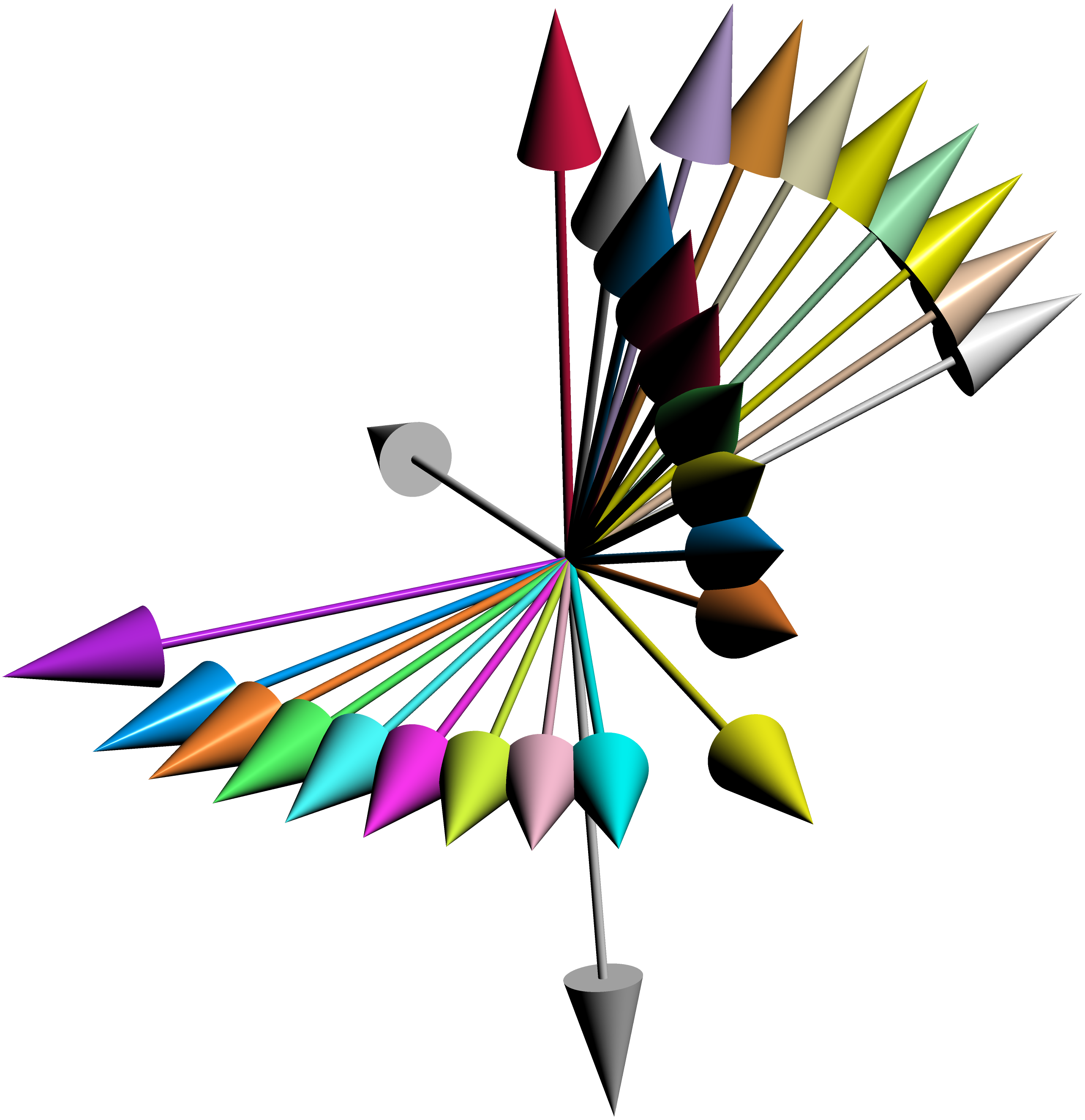}
		\caption{first label set with $29$~labels distributed non-uniformly}
    \label{figure:fandisk-mesh:labels:nonuniform}
  \end{subfigure}
	\hfill
  \begin{subfigure}{0.45\textwidth}
    \centering
    \includegraphics[width = 0.8\textwidth]{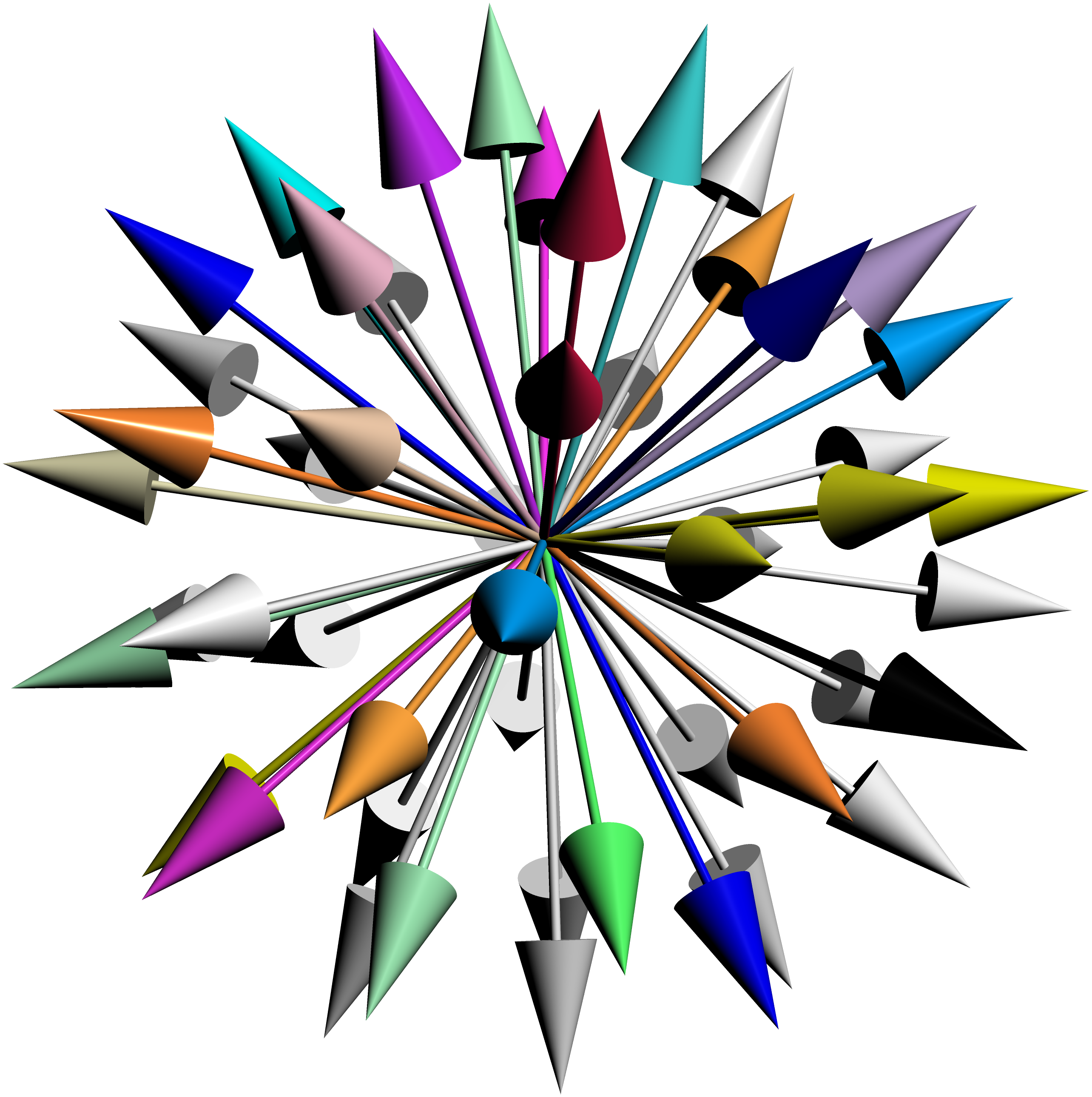}
		\caption{second label set with $50$~labels distributed uniformly}
    \label{figure:fandisk-mesh:labels:uniform}
  \end{subfigure}
  \caption{%
		Visualization of the two label sets for the fandisk mesh example (\cref{subsection:numerical-examples:fandisk-mesh}).
    Labels that are not visible in the corresponding result plots~\cref{figure:fandisk-mesh:nonuniform:atv-ltv,figure:fandisk-mesh:uniform:atv-ltv} are colored in white.
  }
	\label{figure:fandisk-mesh:labels}
\end{figure}

Next, we consider the fandisk mesh from \cite{HoppeDeRoseDuchampHalsteadJinMcDonaldSchweitzerStuetzle:1994:1}, which is available at the Wolfram Data Respository~\cite{Shedelbower:2022:1}.
As above, we added Gaussian noise to the vertex positions.
For this example, we consider two different label sets.
The first label set has $29$~labels and they are chosen to represent the normals occurring on the round parts of the fandisk mesh well.
The aim is to allow for a finer segmentation of the mesh on said parts.
By contrast, the second label set consists of $50$~labels that are distributed uniformly on the sphere.
We obtain this label set using a Fibonacci lattice; see \cite{Gonzalez:2009:1}.
The label sets are visualized in \cref{figure:fandisk-mesh:labels}.

\begin{figure}[htb]
	\centering
	\begin{subfigure}{0.31\textwidth}
		\begin{center}
			\makeatletter
			\ltx@ifclassloaded{amsart}{%
				\includegraphics[width = 0.9\textwidth]{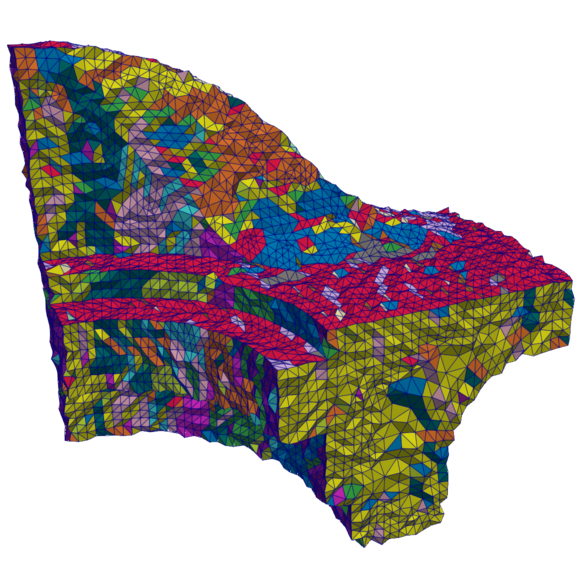}
			}{%
				\includegraphics[width = 0.9\textwidth]{pictures/fandisk_nonuniform_assignmentTV_0_004.png}
			}
			\makeatother
		\end{center}
		\caption{$\TVweight = 0.004$}
		\label{figure:fandisk-mesh:nonuniform:atv:small}
	\end{subfigure}
	\begin{subfigure}{0.31\textwidth}
		\begin{center}
			\makeatletter
			\ltx@ifclassloaded{amsart}{%
				\includegraphics[width = 0.9\textwidth]{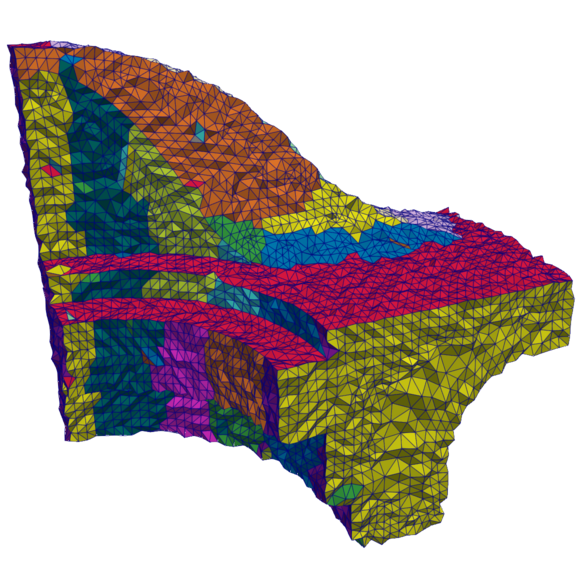}
			}{%
				\includegraphics[width = 0.9\textwidth]{pictures/fandisk_nonuniform_assignmentTV_0_008.png}
			}
			\makeatother
		\end{center}
		\caption{$\TVweight = \TVweight^*_\atv = 0.008$}
		\label{figure:fandisk-mesh:nonuniform:atv:optimal}
	\end{subfigure}
	\begin{subfigure}{0.31\textwidth}
		\begin{center}
			\makeatletter
			\ltx@ifclassloaded{amsart}{%
				\includegraphics[width = 0.9\textwidth]{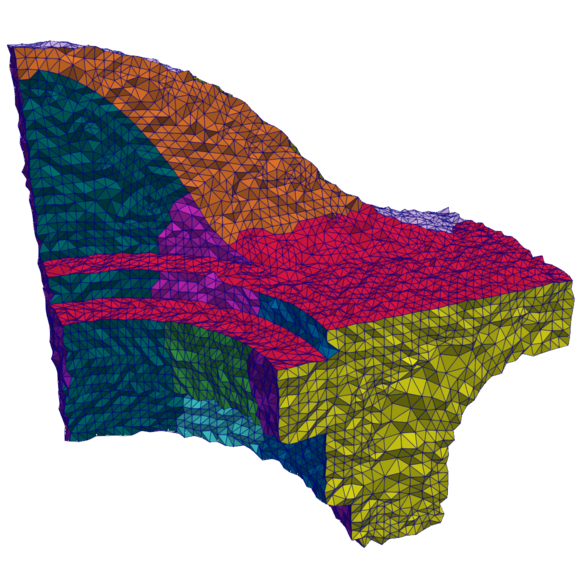}
			}{%
				\includegraphics[width = 0.9\textwidth]{pictures/fandisk_nonuniform_assignmentTV_0_016.png}
			}
			\makeatother
		\end{center}
		\caption{$\TVweight = 0.016$}
		\label{figure:fandisk-mesh:nonuniform:atv:large}
	\end{subfigure}
	\\[\baselineskip]
	\begin{subfigure}{0.31\textwidth}
		\begin{center}
			\makeatletter
			\ltx@ifclassloaded{amsart}{%
				\includegraphics[width = 0.9\textwidth]{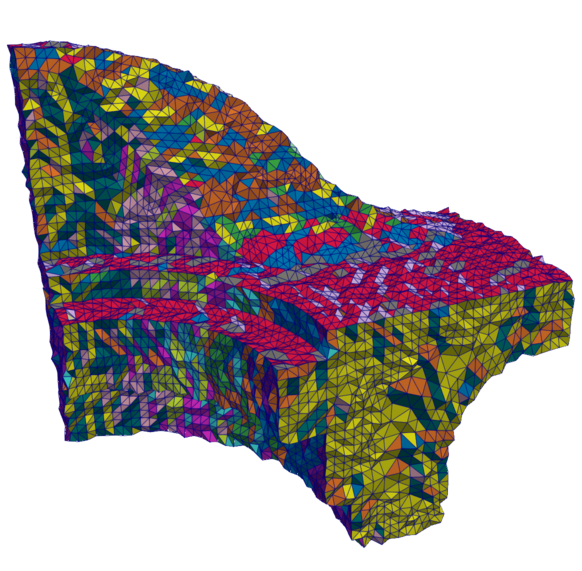}
			}{%
				\includegraphics[width = 0.9\textwidth]{pictures/fandisk_nonuniform_labelTV_0_02.png}
			}
			\makeatother
		\end{center}
		\caption{$\TVweight = 0.02$}
		\label{figure:fandisk-mesh:nonuniform:ltv:small}
	\end{subfigure}
	\begin{subfigure}{0.31\textwidth}
		\begin{center}
			\makeatletter
			\ltx@ifclassloaded{amsart}{%
				\includegraphics[width = 0.9\textwidth]{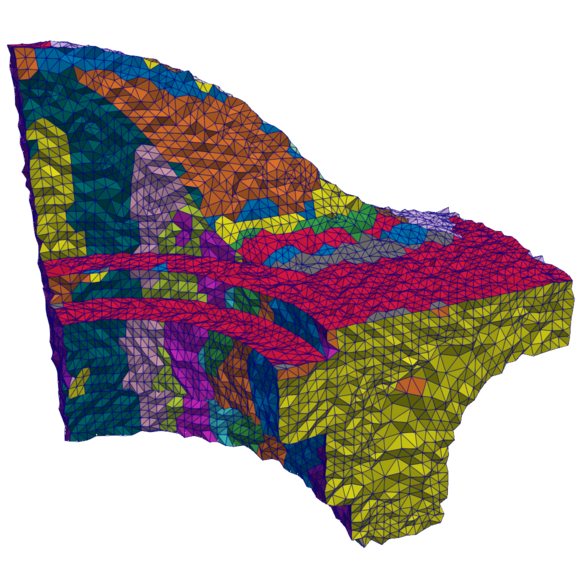}
			}{%
				\includegraphics[width = 0.9\textwidth]{pictures/fandisk_nonuniform_labelTV_0_04.png}
			}
			\makeatother
		\end{center}
		\caption{$\TVweight = \TVweight^*_\ltv = 0.04$}
		\label{figure:fandisk-mesh:nonuniform:ltv:optimal}
	\end{subfigure}
	\begin{subfigure}{0.31\textwidth}
		\begin{center}
			\makeatletter
			\ltx@ifclassloaded{amsart}{%
				\includegraphics[width = 0.9\textwidth]{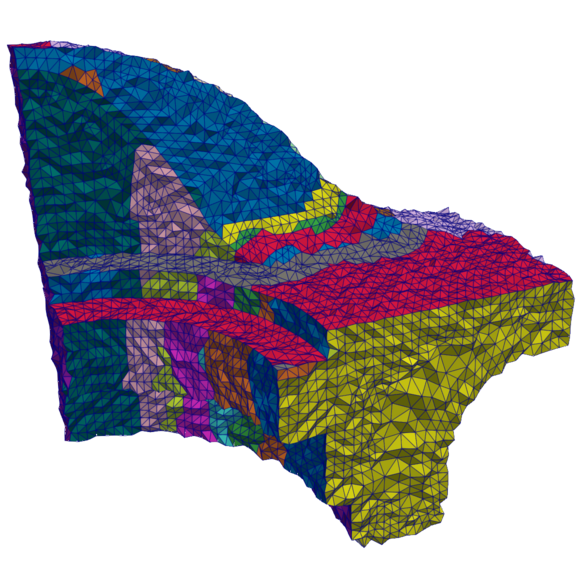}
			}{%
				\includegraphics[width = 0.9\textwidth]{pictures/fandisk_nonuniform_labelTV_0_08.png}
			}
			\makeatother
		\end{center}
		\caption{$\TVweight = 0.08$}
		\label{figure:fandisk-mesh:nonuniform:ltv:large}
	\end{subfigure}
	\caption{%
		Assignments for the noisy fandisk mesh with the first label set of $29$~labels (\cref{figure:fandisk-mesh:labels:nonuniform}) and different values of $\TVweight$.
		Triangles are colored according to their assigned label.
		\Cref{figure:fandisk-mesh:nonuniform:atv:small,figure:fandisk-mesh:nonuniform:atv:optimal,figure:fandisk-mesh:nonuniform:atv:large} show the results using \eqref{eq:atv:model-problem}, while \cref{figure:fandisk-mesh:nonuniform:ltv:small,figure:fandisk-mesh:nonuniform:ltv:optimal,figure:fandisk-mesh:nonuniform:ltv:large} show the results using \eqref{eq:ltv:model-problem}.
		Quantitative results for these examples are given in \cref{table:fandisk-mesh:nonuniform:atv-ltv}.
	}
	\label{figure:fandisk-mesh:nonuniform:atv-ltv}
\end{figure}

The results for the first label set are shown in \cref{figure:fandisk-mesh:nonuniform:atv-ltv}.
For each of the two models, we choose $\beta \in \set{1/2, 1, 2} \cdot \TVweight^*$, where $\TVweight^*$ is the optimal parameter choice for the respective model.

For the optimal parameter~$\TVweight^*_\atv$, the solution of the \eqref{eq:atv:model-problem} model uses 26 out of the available $29$~labels and correctly labels \SI{89.6}{\percent} of the triangles with a Rand index of $0.989$.
The new \eqref{eq:ltv:model-problem} uses $27$~labels and achieves a correctness of \SI{91.4}{\percent} with a Rand index of $0.987$.
Further quantitative results are shown in \cref{table:fandisk-mesh:nonuniform:atv-ltv}.

\begin{table}[htb]
	\robustify\bfseries
  \centering
  \begin{tabular}{
			l
			l
			S[table-format = 1.3, table-model-setup = \bfseries]
			S[table-format = 2]
			S[table-format = 2.1, table-model-setup = \bfseries]
			S[table-format = 1.3, table-model-setup = \bfseries]
		}
    \toprule
    &
    model
    &
    {TV weight $\TVweight$}
    &
    {labels used}
    &
		{correctly labeled~\unit{\percent}}
    &
    {RI}
    \\
    \midrule
    \cref{figure:fandisk-mesh:nonuniform:atv:small}
    &
    \eqref{eq:atv:model-problem}
    &
    0.004
    &
    29
    &
    81.7
    &
    0.975
    \\
    \cref{figure:fandisk-mesh:nonuniform:atv:optimal}
    &
    \eqref{eq:atv:model-problem}
    &
    \bfseries 0.008
    &
    26
    &
    \bfseries 89.6
    &
    \bfseries 0.989
    \\
    \cref{figure:fandisk-mesh:nonuniform:atv:large}
    &
    \eqref{eq:atv:model-problem}
    &
    0.016
    &
    15
    &
    85.5
    &
    0.981
    \\
		\midrule
    \cref{figure:fandisk-mesh:nonuniform:ltv:small}
    &
    \eqref{eq:ltv:model-problem}
    &
    0.02
    &
    29
    &
    81.9
    &
    0.976
    \\
    \cref{figure:fandisk-mesh:nonuniform:ltv:optimal}
    &
    \eqref{eq:ltv:model-problem}
    &
    \bfseries 0.04
    &
    27
    &
    \bfseries 91.4
    &
    \bfseries 0.987
    \\
    \cref{figure:fandisk-mesh:nonuniform:ltv:large}
    &
    \eqref{eq:ltv:model-problem}
    &
    0.08
    &
    27
    &
    85.4
    &
    0.980
    \\
    \bottomrule
  \end{tabular}
	\caption{%
    Quantitative results for the noisy fandisk mesh example (\cref{figure:fandisk-mesh:nonuniform:atv-ltv}) with the first label set $29$~labels for different values of the regularization parameter~$\TVweight$ and the two models.
    The number of correctly labeled triangles and the Rand index are measured with respect to the ground-truth segmentation given by the original mesh without noise and without regularization ($\TVweight = 0$).
  }
  \label{table:fandisk-mesh:nonuniform:atv-ltv}
\end{table}
We observe that the \eqref{eq:ltv:model-problem} is able to assign a more detailed segmentation of the round parts in the front both for the optimal parameter~$\TVweight_\ltv^*$ and also for $2 \, \TVweight_\ltv^*$.
This shows that for this setup, the \eqref{eq:ltv:model-problem} is again more robust against choosing a regularization parameter~$\TVweight$ that is too large.
Both models fail to produce a meaningful segmentation when the regularization parameter is $\TVweight^*/2$.

\begin{figure}[htb]
	\centering
	\begin{subfigure}{0.31\textwidth}
		\begin{center}
			\makeatletter
			\ltx@ifclassloaded{amsart}{%
				\includegraphics[width = 0.9\textwidth]{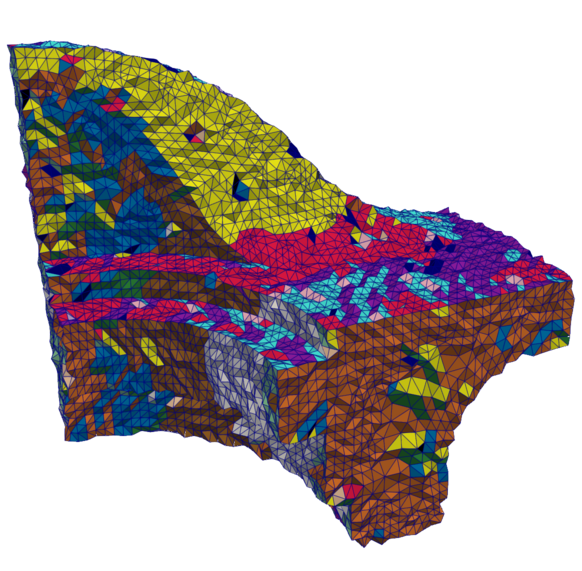}
			}{%
				\includegraphics[width = 0.9\textwidth]{pictures/fandisk_50_assignmentTV_0_008.png}
			}
			\makeatother
		\end{center}
		\caption{$\TVweight = 0.008$}
		\label{figure:fandisk-mesh:uniform:atv:small}
	\end{subfigure}
	\begin{subfigure}{0.31\textwidth}
		\begin{center}
			\makeatletter
			\ltx@ifpackageloaded{amsart}{%
				\includegraphics[width = 0.9\textwidth]{pictures-small/fandisk_50_assignmentTV_0_016.png}
			}{%
				\includegraphics[width = 0.9\textwidth]{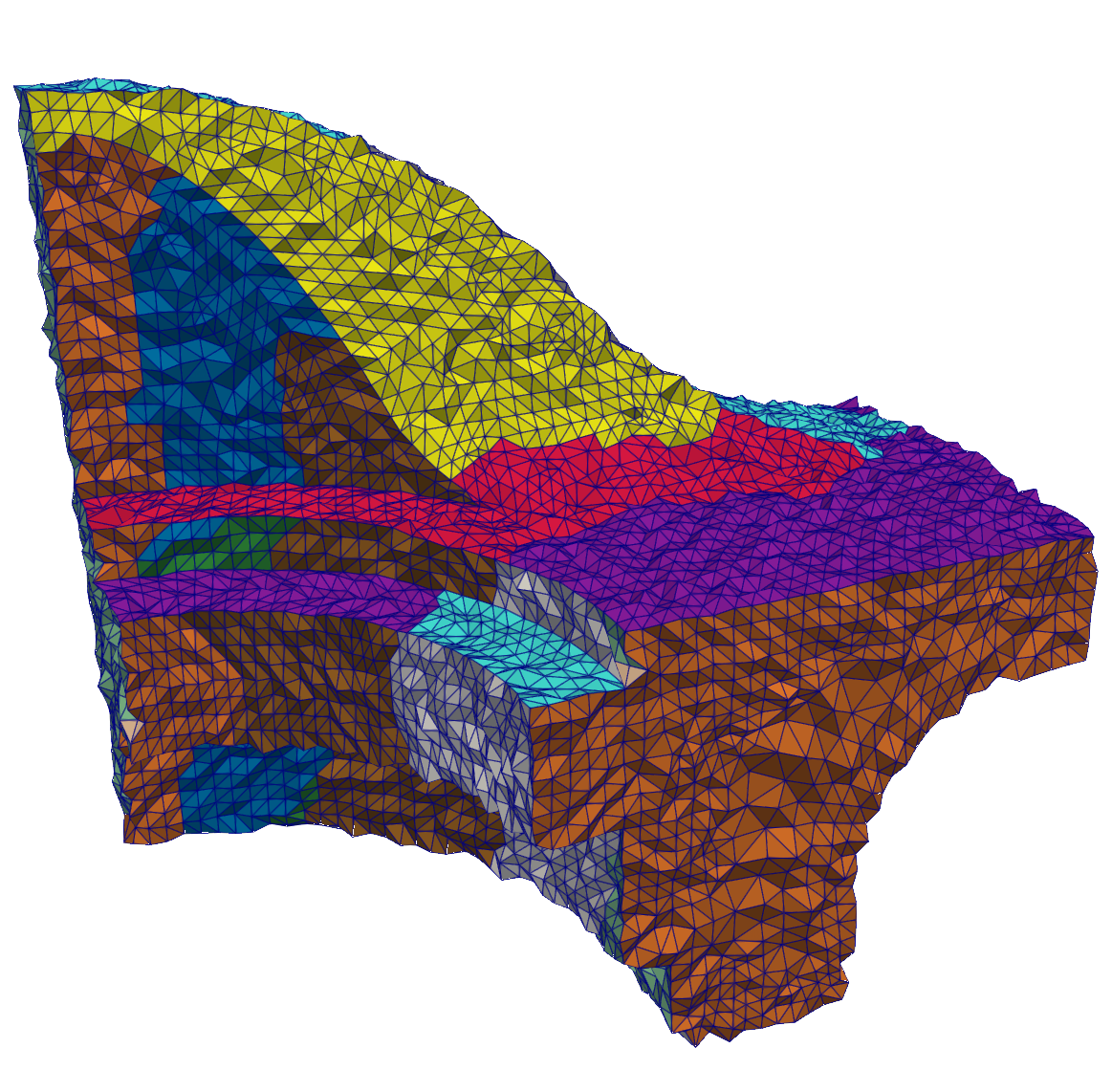}
			}
			\makeatother
		\end{center}
		\caption{$\TVweight = \TVweight^*_\atv = 0.016$}
		\label{figure:fandisk-mesh:uniform:atv:optimal}
	\end{subfigure}
	\begin{subfigure}{0.31\textwidth}
		\begin{center}
			\makeatletter
			\ltx@ifpackageloaded{amsart}{%
				\includegraphics[width = 0.9\textwidth]{pictures-small/fandisk_50_assignmentTV_0_032.png}
			}{%
				\includegraphics[width = 0.9\textwidth]{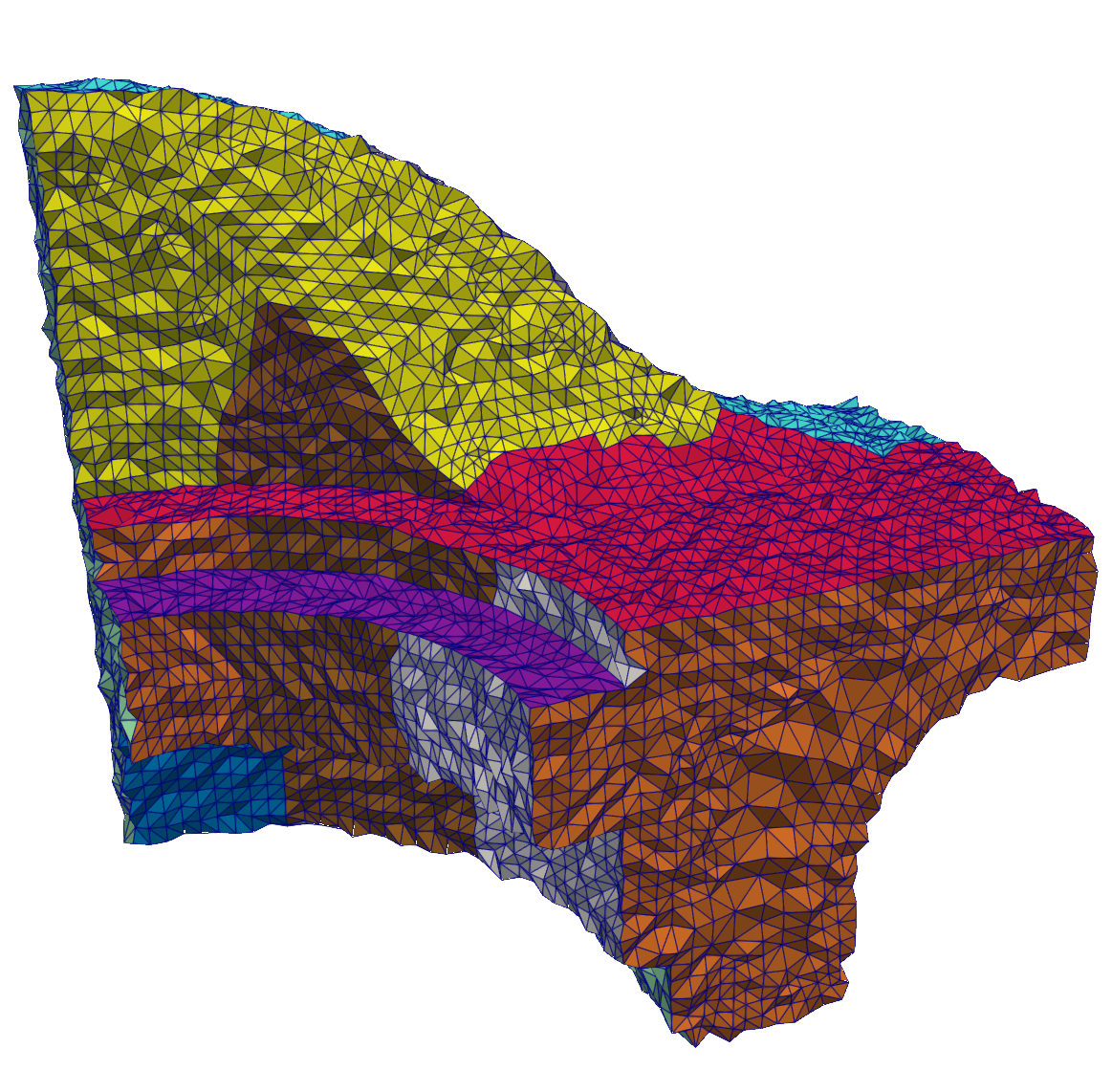}
			}
			\makeatother
		\end{center}
		\caption{$\TVweight = 0.032$}
		\label{figure:fandisk-mesh:uniform:atv:large}
	\end{subfigure}
	\\[\baselineskip]
	\begin{subfigure}{0.31\textwidth}
		\begin{center}
			\makeatletter
			\ltx@ifpackageloaded{amsart}{%
				\includegraphics[width = 0.9\textwidth]{pictures-small/fandisk_50_labelTV_0_025.png}
			}{%
				\includegraphics[width = 0.9\textwidth]{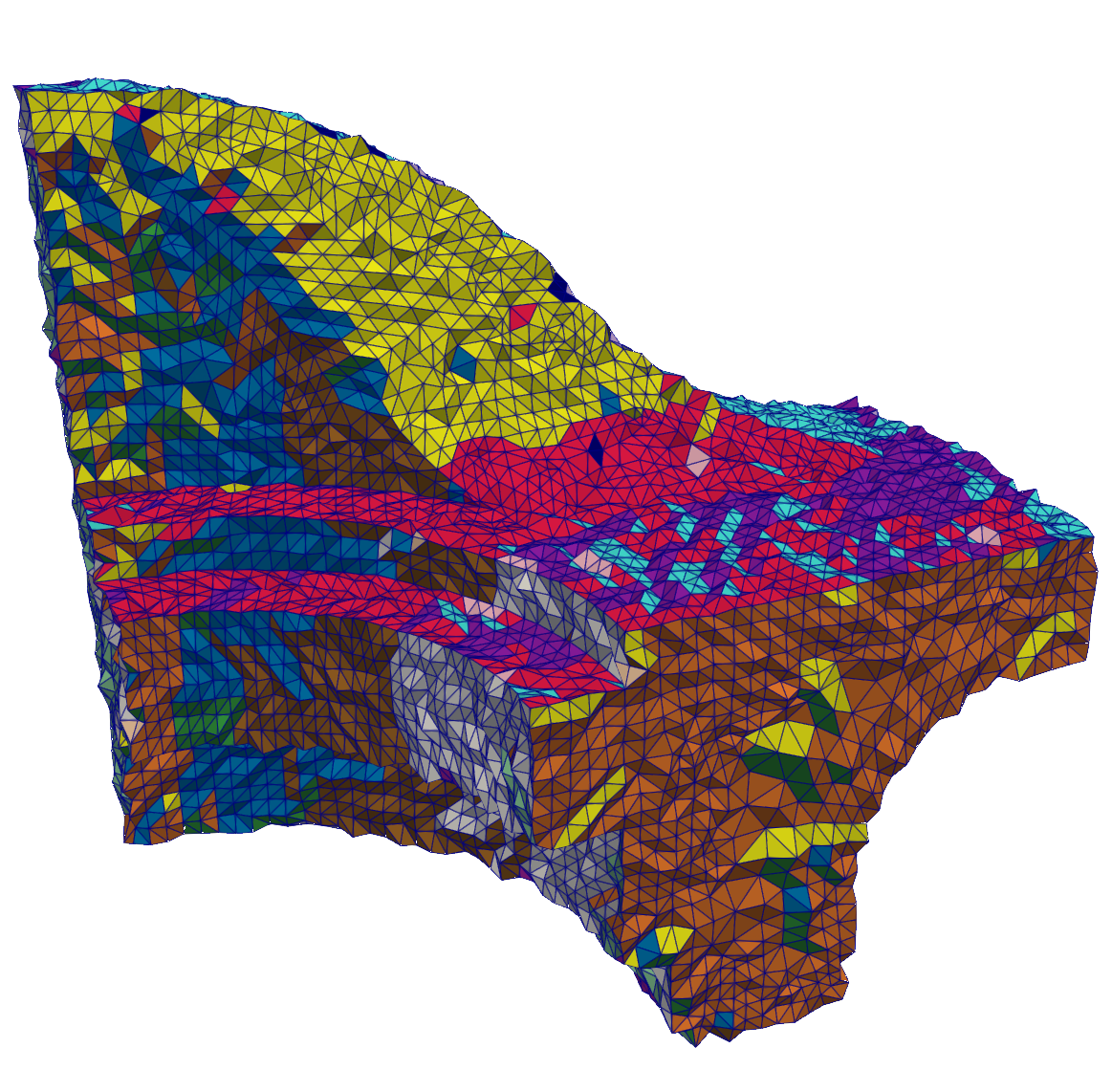}
			}
			\makeatother
		\end{center}
		\caption{$\TVweight = 0.025$}
		\label{figure:fandisk-mesh:uniform:ltv:small}
	\end{subfigure}
	\begin{subfigure}{0.31\textwidth}
		\begin{center}
			\makeatletter
			\ltx@ifpackageloaded{amsart}{%
				\includegraphics[width = 0.9\textwidth]{pictures-small/fandisk_50_labelTV_0_05.png}
			}{%
				\includegraphics[width = 0.9\textwidth]{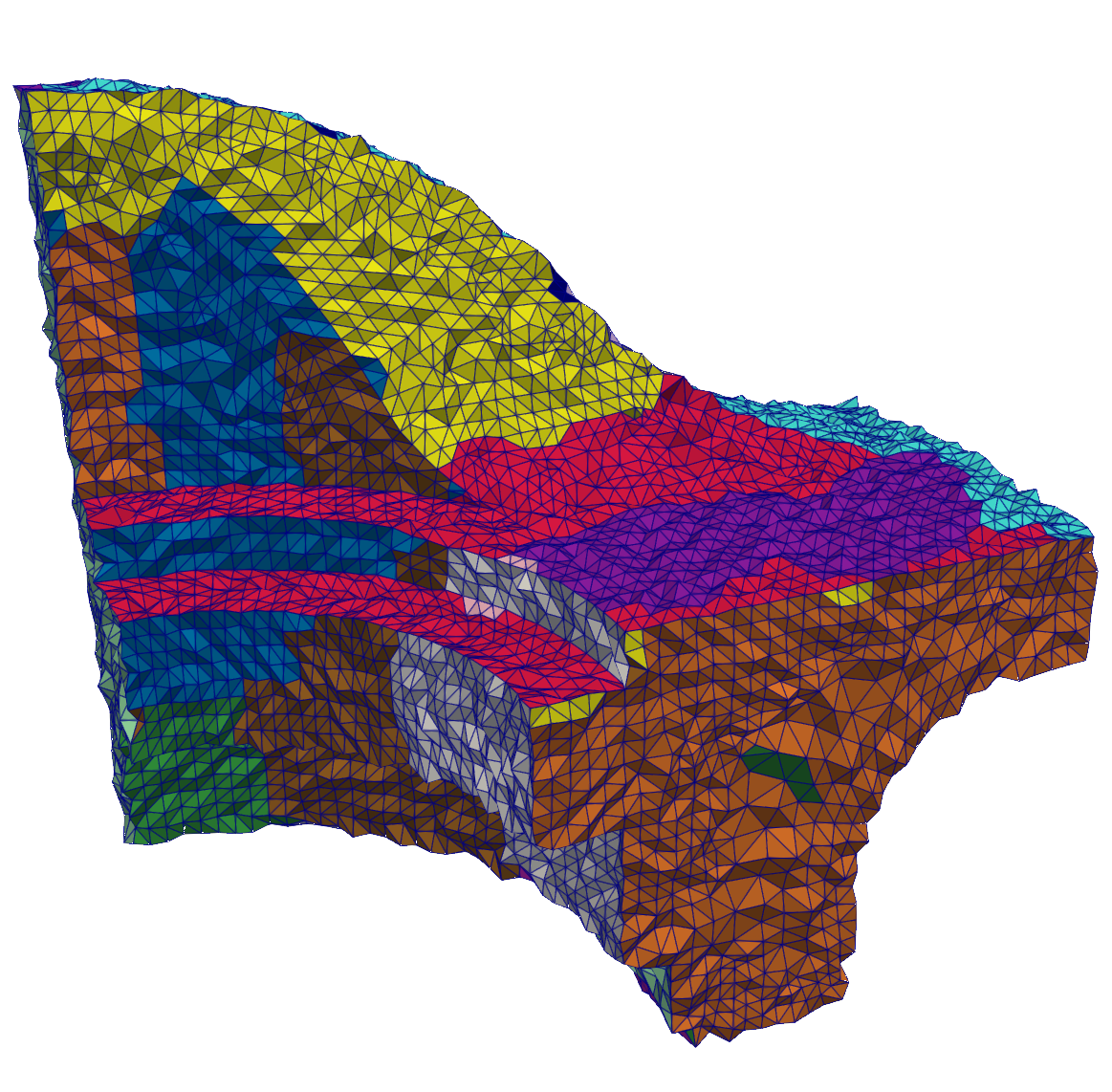}
			}
			\makeatother
		\end{center}
		\caption{$\TVweight = \TVweight^*_\ltv = 0.05$}
		\label{figure:fandisk-mesh:uniform:ltv:optimal}
	\end{subfigure}
	\begin{subfigure}{0.31\textwidth}
		\begin{center}
			\makeatletter
			\ltx@ifpackageloaded{amsart}{%
				\includegraphics[width = 0.9\textwidth]{pictures-small/fandisk_50_labelTV_0_1.png}
			}{%
				\includegraphics[width = 0.9\textwidth]{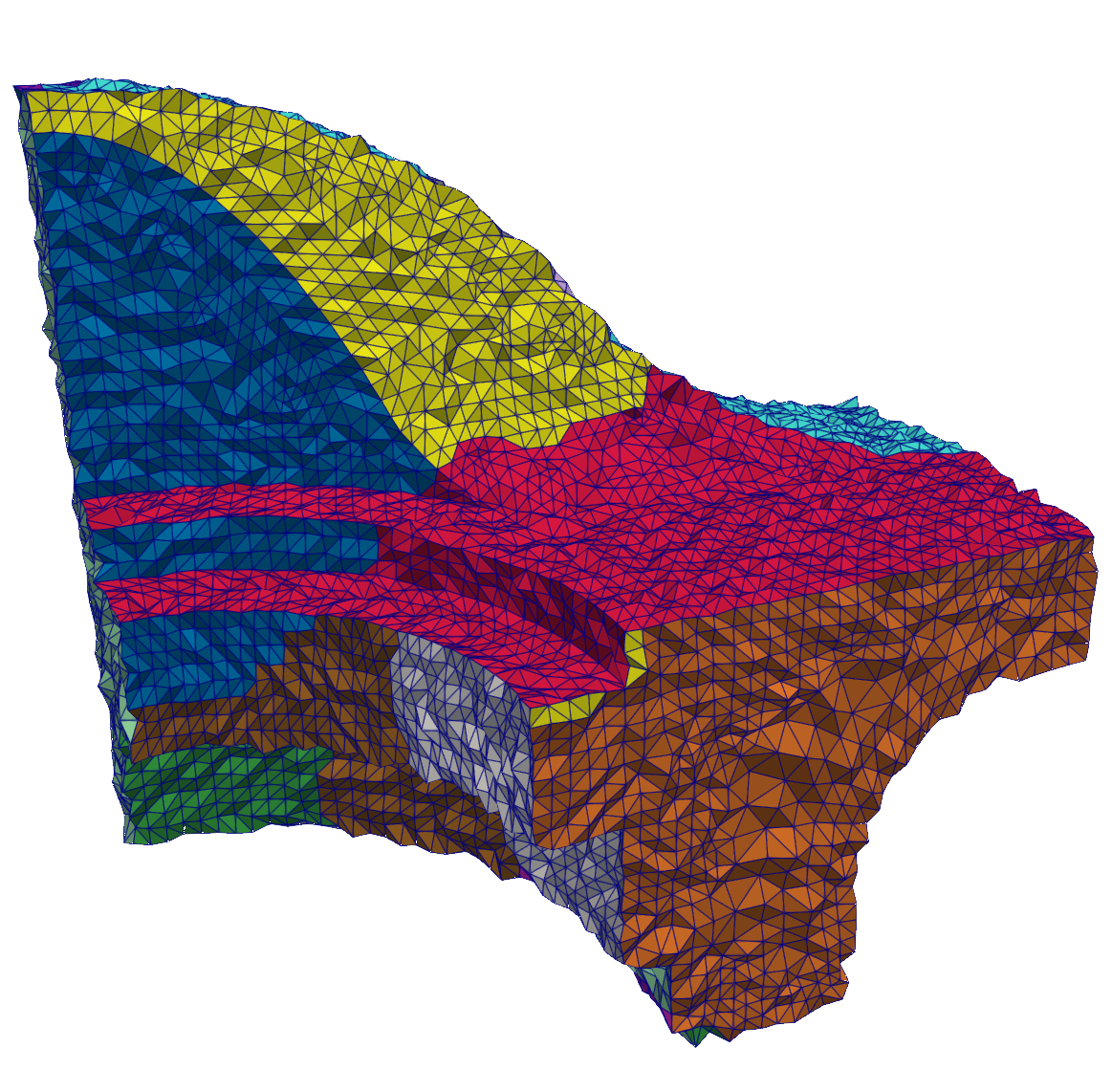}
			}
			\makeatother
		\end{center}
		\caption{$\TVweight = 0.1$}
		\label{figure:fandisk-mesh:uniform:ltv:large}
	\end{subfigure}
	\caption{%
		Assignments for the noisy fandisk mesh with the second label set of $50$~uniformly distributed labels (\cref{figure:fandisk-mesh:labels:uniform}) with different values of $\TVweight$.
		Triangles are colored according to their assigned label.
		\Cref{figure:fandisk-mesh:uniform:atv:small,figure:fandisk-mesh:uniform:atv:optimal,figure:fandisk-mesh:uniform:atv:large} show the results using \eqref{eq:atv:model-problem}, while \cref{figure:fandisk-mesh:uniform:ltv:small,figure:fandisk-mesh:uniform:ltv:optimal,figure:fandisk-mesh:uniform:ltv:large} show the results using \eqref{eq:ltv:model-problem}.
		Quantitative results for these examples are given in \cref{table:fandisk-mesh:uniform:atv-ltv}.
	}
	\label{figure:fandisk-mesh:uniform:atv-ltv}
\end{figure}

The results for the second label set are shown in \cref{figure:fandisk-mesh:uniform:atv-ltv}.

In this case, the \eqref{eq:atv:model-problem} model uses $26$ of the available $50$~labels and correctly labels $\SI{86.8}{\percent}$ of the mesh with a Rand index of $0.978$, while the \eqref{eq:ltv:model-problem} model uses $35$~labels and correctly labels $\SI{85.2}{\percent}$ of the mesh with a Rand index of $0.977$.
Further quantitative results are shown in \cref{table:fandisk-mesh:uniform:atv-ltv}.

\begin{table}[htb]
	\robustify\bfseries
  \centering
  \begin{tabular}{
			l
			l
			S[table-format = 1.3, table-model-setup = \bfseries]
			S[table-format = 2]
			S[table-format = 2.1, table-model-setup = \bfseries]
			S[table-format = 1.3, table-model-setup = \bfseries]
		}
    \toprule
    &
    model
    &
    {TV weight $\TVweight$}
    &
    {labels used}
    &
    {correctly labeled~\unit{\percent}}
    &
    {RI}
    \\
    \midrule
    \cref{figure:fandisk-mesh:uniform:atv:small}
    &
    \eqref{eq:atv:model-problem}
    &
    0.008
    &
    50
    &
    73.3
    &
    0.955
    \\
    \cref{figure:fandisk-mesh:uniform:atv:optimal}
    &
    \eqref{eq:atv:model-problem}
    &
    \bfseries 0.016
    &
    26
    &
    \bfseries 86.8
    &
    \bfseries 0.978
    \\
    \cref{figure:fandisk-mesh:uniform:atv:large}
    &
    \eqref{eq:atv:model-problem}
    &
    0.032
    &
    15
    &
    79.6
    &
    0.969
    \\
		\midrule
    \cref{figure:fandisk-mesh:uniform:ltv:small}
    &
    \eqref{eq:ltv:model-problem}
    &
    0.025
    &
    46
    &
    75.3
    &
    0.957
    \\
    \cref{figure:fandisk-mesh:uniform:ltv:optimal}
    &
    \eqref{eq:ltv:model-problem}
    &
    \bfseries 0.05
    &
    35
    &
    \bfseries 85.2
    &
    \bfseries 0.977
    \\
    \cref{figure:fandisk-mesh:uniform:ltv:large}
    &
    \eqref{eq:ltv:model-problem}
    &
    0.1
    &
    32
    &
    74.3
    &
    0.959
    \\
    \bottomrule
  \end{tabular}
	\caption{%
    Quantitative results for the noisy fandisk mesh example (\cref{figure:fandisk-mesh:uniform:atv-ltv}) with the second label set of $50$~uniformly distributed labels for different values of the regularization parameter~$\TVweight$ and the two models.
    The number of correctly labeled triangles and the Rand index are measured with respect to the ground-truth segmentation given by the original mesh without noise and without regularization ($\TVweight = 0$).
  }
  \label{table:fandisk-mesh:uniform:atv-ltv}
\end{table}
We see that \eqref{eq:ltv:model-problem} is having difficulties in removing the noise near the sharp boundaries between labels, even for higher values of the regularization parameter~$\TVweight$.
On the other hand, the \eqref{eq:ltv:model-problem} is able to more precisely recover the round parts of the original mesh and remove the noise.

\section{Conclusion}
\label{section:conclusion}

In this paper, we have compared two variational models for the segmentation of triangulated surfaces based on the normal vector field as the governing feature.
The classical assignment space total variation model \eqref{eq:atv:model-problem} penalizes the total variation of the assignment function directly.
Every transition from one vertex in the assignment simplex to another is equally penalized with a value of~$2$.
By contrast, the label space total variation model \eqref{eq:ltv:model-problem} penalizes the total variation of the normal vector represented by the assignment function in the label space~$\Sphere$.
Consequently, every transition from one vertex in the assignment simplex to another has a cost proportional to the geodesic distance between the corresponding labels on the sphere.

While \eqref{eq:ltv:model-problem} appears equally natural, it is computationally significantly more challenging to solve than \eqref{eq:atv:model-problem}.
This additional complexity needs to be taken into account when selecting between the two models for a given application.
The added complexity is due to the fact that the labels are situated in a nonlinear space (the sphere), and the evaluation of their mixture requires a Riemannian center of mass problem to be solved on each triangle as part of the optimization process.
In order to mitigate the computational overhead, we proposed a novel manifold Newton method for the subproblem arising from the Riemannian center of mass problem in the ADMM loop.
Numerical experiments show that this speeds up the overall optimization significantly compared to a gradient descent approach.
Still, the \eqref{eq:ltv:model-problem} remains computationally more expensive than \eqref{eq:atv:model-problem} when solved using ADMM.

We observed that \eqref{eq:atv:model-problem} does not necessarily use all available labels, especially for large values of the regularization parameter.
In particular, this can be observed in regions of constant curvature and multiple labels available for this region.
In contrast, the proposed model \eqref{eq:ltv:model-problem} utilizes more of the available labels.
It also turns out to be more robust \wrt the choice of the regularization parameter~$\TVweight$.

\appendix

\printbibliography

\end{document}